%% file: main.tex
\documentclass[twoside]{article}

\usepackage[accepted]{aistats2026}

\usepackage{microtype}
\usepackage{graphicx}
\usepackage{booktabs} %
\usepackage{algorithm}
\usepackage{algorithmic}
\usepackage{wrapfig}
\usepackage{hyperref}
\usepackage{multicol}

\usepackage{amsmath}
\usepackage{amssymb}
\usepackage{mathtools}
\usepackage{amsthm}
\usepackage[textsize=tiny]{todonotes}
\usepackage{bbm}
\usepackage{placeins}
\usepackage{multirow}
\usepackage{subcaption}
\usepackage[capitalize,noabbrev]{cleveref}
\usepackage{minitoc}

\theoremstyle{plain}
\newtheorem{theorem}{Theorem}[section]
\newtheorem{proposition}[theorem]{Proposition}
\newtheorem{lemma}[theorem]{Lemma}

\theoremstyle{definition}
\newtheorem{definition}[theorem]{Definition}

\theoremstyle{remark}

\definecolor{charteGreen}{RGB}{80,190,135}

\input{math_commands}

\runningtitle{An Indicator of MIS in Post-Training Quantized Models}

\begin{document}

\doparttoc %
\faketableofcontents %

\twocolumn[
\aistatstitle{An Indicator of Membership Inference Security in \\ Post-Training Quantized Models}
\aistatsauthor{
Eric Aubinais$^{1,*}$\And Philippe Formont$^{2,*}$ \And Pablo Piantanida$^3$ \And Elisabeth Gassiat$^1$
}
\aistatsaddress{
$^{1}$Universit\'e Paris-Saclay, CNRS, Laboratoire de mathématiques d’Orsay, France \\
$^{2}$Universit\'e Paris-Saclay, ILLS, MILA, ÉTS, Montreal, Canada \\
$^{3}$ILLS, MILA, CNRS, CentraleSupélec, Montreal, Canada
}
]

\input{macros}

\begin{abstract}

    \input{Sections/Abstract}
\end{abstract}
\vspace{0.3cm}
{\footnotesize $^*$ Equal contribution.}

\section{INTRODUCTION}
\label{sec:intro}

\input{Sections/I-Introduction}

\section{BACKGROUND AND NOTATIONS}
\label{sec:context}
\input{Sections/II-Context}

\section{ATTACK-FREE MIS EVALUATION FOR QUANTIZED PROCEDURES}
\label{sec:theo_res}
\input{Sections/III-Theoretical_Results}

\subsection{Estimation of the MIS Indicator}
\label{sec:algo}
\input{Sections/IV-Algorithm}

\input{Sections/V-Numerical_Experiments}

\section{CONCLUSION, LIMITATIONS AND PERSPECTIVES}
\label{sec:limitations}
\input{Sections/VI-Limitations}

\clearpage

\bibliography{bibliography}

\section*{Checklist}

\begin{enumerate}

  \item For all models and algorithms presented, check if you include:
  \begin{enumerate}
    \item A clear description of the mathematical setting, assumptions, algorithm, and/or model. [Yes]
    \item An analysis of the properties and complexity (time, space, sample size) of any algorithm. [Yes]
    \item (Optional) Anonymized source code, with specification of all dependencies, including external libraries. [Yes]
  \end{enumerate}

  \item For any theoretical claim, check if you include:
  \begin{enumerate}
    \item Statements of the full set of assumptions of all theoretical results. [Yes]
    \item Complete proofs of all theoretical results. [Yes]
    \item Clear explanations of any assumptions. [Yes]     
  \end{enumerate}

  \item For all figures and tables that present empirical results, check if you include:
  \begin{enumerate}
    \item The code, data, and instructions needed to reproduce the main experimental results (either in the supplemental material or as a URL). [Yes]
    \item All the training details (e.g., data splits, hyperparameters, how they were chosen). [Yes]
    \item A clear definition of the specific measure or statistics and error bars (e.g., with respect to the random seed after running experiments multiple times). [Yes]
    \item A description of the computing infrastructure used. (e.g., type of GPUs, internal cluster, or cloud provider). [Yes]
  \end{enumerate}

  \item If you are using existing assets (e.g., code, data, models) or curating/releasing new assets, check if you include:
  \begin{enumerate}
    \item Citations of the creator If your work uses existing assets. [Yes]
    \item The license information of the assets, if applicable. [Not Applicable]
    \item New assets either in the supplemental material or as a URL, if applicable. [Not Applicable]
    \item Information about consent from data providers/curators. [Not Applicable]
    \item Discussion of sensible content if applicable, e.g., personally identifiable information or offensive content. [Not Applicable]
  \end{enumerate}

  \item If you used crowdsourcing or conducted research with human subjects, check if you include:
  \begin{enumerate}
    \item The full text of instructions given to participants and screenshots. [Not Applicable]
    \item Descriptions of potential participant risks, with links to Institutional Review Board (IRB) approvals if applicable. [Not Applicable]
    \item The estimated hourly wage paid to participants and the total amount spent on participant compensation. [Not Applicable]
  \end{enumerate}

\end{enumerate}

\newpage
\onecolumn
\aistatstitle{An Indicator of Membership Inference Security in Post-Training Quantized Models: 
Supplementary Materials}

\part{Supplementary Materials}
\addcontentsline{toc}{section}{Supplementary Materials}
\parttoc

\newpage
\input{Appendix/XVI-Table_of_Notations}

\input{Appendix/XI-A}

\input{Appendix/XII-Quantizations}
\input{Appendix/XIII-Synthetic_Details}

\input{Appendix/XIV-NLP_Details}

\input{Appendix/XV-Molecular_Details}

\end{document}

%% file: math_commands.tex
\usepackage{amsmath}
\usepackage{amssymb, bm}
\usepackage{mathtools}

\def\ccref#1#2{\hyperref[#2]{#1 \ref*{#2}}}

\newtheorem{example}[theorem]{Example}

\def\toinf#1{\underset{#1\to\infty}{\to}}
\def\liminf#1{\underset{#1\to\infty}{\lim}\;}
\def\pp#1{\left(#1\right)}

\def\cc#1{\left[#1\right]}

\def\eqref#1{equation~\ref{#1}}

\def\1{\bm{1}}

\def\rx{{\textnormal{x}}}

\def\rz{{\textnormal{z}}}

\def\gA{{\mathcal{A}}}

\def\gD{{\mathcal{D}}}

\def\gF{{\mathcal{F}}}

\def\gL{{\mathcal{L}}}

\def\gN{{\mathcal{N}}}

\def\gQ{{\mathcal{Q}}}
\def\gR{{\mathcal{R}}}

\def\gZ{{\mathcal{Z}}}

\newcommand{\E}{\mathbb{E}}
\newcommand{\Ls}{\mathcal{L}}
\newcommand{\R}{\mathbb{R}}

\DeclareMathOperator*{\argmin}{arg\,min}

%% file: macros.tex
\def\bsl{g_\psi}
\def\Acc{\texttt{Acc}}
\def\MIS{\texttt{MIS}}
\def\Deltan{\Delta_n}
\def\nrunmol{10 }
\def\thetan{{\hat{\theta}_n}}
\def\thetaq{{\Bar{\theta}}}
\def\Thetaq{{\Bar{\Theta}}}

\newcommand{\mkn}[1]{\delta_{#1}^n}
\newcommand{\diffmetric}[2]{r^{#1}_{#2}}

\def\Deltagen{\Delta_{\eta,\lambda,n}}

\def\qcertif{r_{\gQ_n}}

\newcommand{\citep}[1]{\cite{#1}}

%% file: Sections/Abstract.tex
Quantizing machine learning models has demonstrated its effectiveness in lowering memory and inference costs while maintaining performance levels comparable to those of the original models.
In this work, we investigate the impact of quantization procedures on privacy in data-driven models, focusing on their vulnerability to membership inference attacks. Membership Inference Security (MIS) has recently been proposed to characterize the privacy of machine learning models against the most powerful (and possibly unknown) attacks. However, quantifying MIS appears to be computationally very difficult. In this paper, we propose a new MIS indicator for post-training quantization procedures of machine learning models that minimizes an empirical loss. This new indicator is a byproduct of a theoretical asymptotic analysis of the MIS in this context. We also present a methodology for empirically estimating our MIS indicator. Using synthetic datasets and real-world data (in the context of drug discovery), we demonstrate the effectiveness of our approach in assessing and ranking the MIS of different quantizers.

%% file: Sections/I-Introduction.tex
Reducing machine learning models' computational and memory costs is a critical aspect of their deployment, particularly on edge devices and resource-constrained environments. Quantization stands out among the various methods available to enhance inference efficiency in neural networks, such as knowledge distillation and pruning, due to its distinct advantages and proven practical success~\citep{gholami2022survey}. One key benefit of quantization is that the storage and latency improvements achieved through quantization are deterministically defined by the chosen quantization level (e.g., using 8-bit integers instead of 32-bit floating-point numbers). Moreover, uniform quantization is inherently hardware-friendly, facilitating the practical realization of theoretical efficiency gains.

While quantization effectively improves efficiency, its impact on the privacy of machine learning models remains largely underexplored. A particularly intriguing question is whether quantization can also strengthen a model's resilience against adversarial threats, such as extracting sensitive information. By reducing the precision of a model's parameters, quantization naturally discards some information~\citep{6451278}, which leads to the hypothesis that this process could potentially reduce the risk of recovering the model's training data or other private information. However, to the best of our knowledge, the security of quantized models against such privacy attacks has not yet been theoretically investigated.

In this paper, we study how quantization affects the vulnerability of machine learning models to Membership Inference Attacks (MIAs). Privacy evaluation has traditionally relied on empirical attack-based benchmarks, which, while informative, are inherently limited by their dependence on specific attack strategies. To address this, \citep{aubinais2023fundamental} introduced Membership Inference Security (MIS), a measure that captures a model's vulnerability to the strongest possible MIA, regardless of the attacker's knowledge. Although MIS is attack-independent, its direct numerical evaluation is computationally demanding, highlighting the need for practical estimation methods. To overcome this difficulty, we propose a novel indicator of MIS and a methodology to estimate it for quantized models trained with empirical risk minimization. Our indicator is grounded in a theoretical asymptotic analysis of MIS as the training dataset size increases, offering both interpretability and practical feasibility.

\subsection{Our contributions}
\label{subsec:contr}
We consider learning procedures minimizing an empirical loss for a loss function $\ell$. Our contributions can be summarized as follows: 
\begin{itemize}
    \item We establish that, for large datasets, the MIS of a learning algorithm under quantization is fully characterized by the distribution of the per-sample loss across quantized models when the model architecture and/or quantization strategy adapt to the training set size~(\autoref{thm:seqQ}).

    Building on this result, we propose a new MIS indicator that is equal to
    $$
   \qcertif\coloneqq \frac{1}{2}\underset{k}{\min}\frac{\cc{\E\cc{\ell(\thetaq^n_k,\rz)}-\E\cc{\ell(\thetaq^n_{k^*},\rz)}}^2}{\textrm{Var}\cc{\ell(\thetaq^n_k,\rz)-\ell(\thetaq^n_{k^*},\rz)}}.
    $$
    Here $n$ is the size of the dataset; $\thetaq^n_k$  are the given quantized parameters, $k^* \in \argmin \E\cc{\ell(\thetaq^n_k,\rz)}$;
    and $\rz$ is a random sample. This indicator enables the comparison of quantization procedures in terms of MIS. 
    \item
    We propose a methodology for estimating this new indicator based on training trajectories. We empirically validate our method for various quantization techniques on synthetic datasets. We show that the rankings induced by our method consistently correlate with the baseline MIS estimations. 
    
    \item On real-world data, we further investigate its relationship to classical MIAs against language models and its connection to downstream performance on molecular prediction tasks (see~\autoref{sec:num_exp} and the supplementary material).
\end{itemize}

Code and experimental materials are available at our anonymous repository:\url{https://anonymous.4open.science/r/Mol_Downstream-B3DB/}.

\subsection{Related work}

\paragraph{Quantization of neural networks.} Several quantization procedures have been studied and employed with the deployment of neural networks on edge devices \citep{yuan2024vit, lin2024awq}, where inference should be time and memory-efficient. Quantization usually answers this task by reducing the (bit-)precision of the parameters of the neural networks, demonstrating effectiveness in Large Language Models \citep{gong2024survey, zhu2024survey} even when the quantization is as strong as 1-bit precision quantization \citep{wang2023bitnet, ma2024fbi}, 1.58-bits precision quantization \citep{ma2024era1bitllmslarge}, arbitrary bits precision \citep{zeng2024abq}.
The most adopted framework of quantization is Post-Training Quantization (PTQ)~\citep{jacob2018quantization, nagel2019data, gholami2022survey}, which provides a simple training-free implementation. PTQ is usually adopted over Quantization-Aware Training (QAT)~\citep{banner2018scalable, pang2024push, nagel2021white, nagel2022overcoming, bengio2013estimating} due to their limitations to scale up to larger models~\citep{gholami2022survey, lin2024awq}. Additionally, some lines of work study "hardware-aware" quantization procedures \citep{wang2024ladder, balaskas2024hardware} where optimization is made directly on the hardware. During our experiments, we will focus on PTQ.

\paragraph{Membership Inference Attacks.}
Membership Inference Attacks (MIAs) can reveal sensitive information~\citep{zarifzadeh2024lowcosthighpowermembershipinference,carlini2022membership,shokri2017membership, song2017machine, carlini2023extracting} about one's data by leveraging the information stored in the parameters of the ML model \citep{hartley2022measuring, del2023bounding}. An extensive line of work has developed in the past decade to construct ever so powerful MIAs in embedding models \citep{song2020information}, regression models \citep{gupta2021membership}, or generative models \citep{hayes202588705membership}, systematically summarized in \citep{hu2022membership}. Recent works have leveraged the predictive power of LLMs to construct new MIAs \citep{staab2023beyond, wang2025survey}. %
Several privacy benchmarks have been developed to evaluate the privacy risks of ML models by testing state-of-the-art MIAs on the target model \citep{murakonda2020ml, liu2022ml}.
While these benchmarks provide valuable information about the privacy leakage of an ML model, specific MIAs cannot fully assess the overall privacy resilience of a learning procedure against various attacks. 
Few works have delved into the theoretical intricacies of MIAs \citep{sablayrolles2019white, del2023bounding}. More recently, \citep{aubinais2023fundamental} derived
the statistical quantity that governs the effectiveness and success of such attacks.

\paragraph{Quantization and Privacy.} Various strategies to protect models from attacks like MIAs have been proposed. In federated learning, the effects of input and gradient quantization have been analyzed through the lens of differential privacy~\citep{youn2023randomizedquantizationneeddifferential, yan_killing_2024, pmlr-v180-chaudhuri22a}. %
The impact of model quantization on security has been primarily assessed through empirical evaluations of MIAs, with no existing theoretical analysis~\citep{kowalski_towards_2022, s23187722}. Our work aims to fill this gap by providing a rigorous theoretical evaluation of the security implications of model quantization.

%% file: Sections/II-Context.tex
\subsection{Notations}
\label{subsec:notations}
Throughout the article, we consider a dataset $\gD_n$ of $n$ independent and identically distributed (i.i.d.) random variables $\rz_1,\cdots,\rz_n$ drawn from a common distribution $P$ over a space $\gZ$. We assume that the goal of the machine learning model is to infer a predictive function $\hat{\Psi}$ from a set of predictors $\gF\coloneqq\{\Psi_\theta: \theta\in\Theta\}$ indexed by some space $\Theta\subseteq\R^d$. We define a \textbf{learning procedure} (algorithm) as a function $\gA:\bigcup_{n\geq1}\gZ^n\to \Theta$. By denoting $\hat{\theta}_n = \gA(\rz_1,\cdots,\rz_n)\in\Theta$, we systematically set $\hat{\Psi} = \Psi_{\hat{\theta}_n}$.
We focus on  learning procedures minimizing an empirical loss $\theta\mapsto\frac{1}{n}\sum_{j=1}^{n}\ell (\theta,\rz_j)$ where $\ell:\Theta\times\gZ\to\R^+$ is the loss function.

Notice that this includes classification tasks, as well as regression or generation tasks.

In the present work, we evaluate the privacy of a learning procedure $\gA$ through Membership Inference Attacks (MIAs). Particularly, in a scenario where $\gD_n$ consists of sensitive data and the model $\hat\Psi$ has been shared (such as a sold product), MIAs pose a notable threat to the privacy of the dataset. MIAs aim at inferring membership of a test sample $\Tilde{\rz}$ to the dataset $\gD_n$ by observing $\hat{\theta}_n$. MIAs can be defined as follows.

\begin{definition}[Membership Inference Attack - MIA] Any measurable map $\phi : \Theta\times\gZ\to\{\text{member},\text{non-member}\}$ is considered to be a \textit{Membership Inference Attack}.
\end{definition}

Throughout the paper, we assume that the MIA may have access to additional information, including the data distribution $P$, the learning procedure $\gA$, and/or other meta-parameters. This framework is usually referred to as white-box \citep{hu2022membership}. Other attack paradigms are included in this setting, such as black-box attacks, which can be considered a restrictive sub-category of white-box attacks.

The remainder of this section introduces important results from \citep{aubinais2023fundamental} that will serve as the foundation for our analysis.

\subsection{MIA accuracy}
\label{subsec:miaacc}

To evaluate the performance of an MIA $\phi$ against a learning procedure $\gA$, we introduce the concept of MIA accuracy, which is defined as a weighted average between the true positive rate (TPR) and the true negative rate (TNR) of the MIA: $\texttt{TNR}(\phi)+\lambda \texttt{TPR}(\phi)$, with 
$\lambda>0$. Note that,  by Lagrange duality, maximizing TPR under small TNR is equivalent to maximizing such a weighted average, in which $\lambda$ is related to the upper bound on  TNR. 

Although individual MIAs can reveal cases of information leakage, they do not offer a complete evaluation of a model's privacy risks. 
 Additionally, we assume no prior knowledge of potential attacks. Therefore, to thoroughly examine the privacy level of a learning procedure, we focus on the highest accuracy that can be achieved over all MIAs, including unknown ones. However, taking the supremum over $\phi$ in $\texttt{TNR}(\phi)+\lambda \texttt{TPR}(\phi)$ 
 leads to a degenerate non-informative result if we aim to compare algorithms.
 As a result, we define the accuracy as follows. 

\begin{definition}[Accuracy of a given MIA] The \textit{accuracy of an MIA} $\phi$ is defined as 
\begin{align}
\begin{split}
    \label{def:perf}
{\Acc}_n(\phi; P,\gA ) &\coloneqq \mathbb{E}[\texttt{TNR}(\phi)] + \lambda\mathbb{E}[\texttt{TPR}(\phi)],
\end{split}
\end{align}
where the expectation is considered over all sources of randomness inherent in the underlying training model and the data used for both training and evaluation. In particular, 
we assume that the test sample $\Tilde{\rz}$ is drawn at random over $\{\rz_1,\cdots,\rz_n\}$ with probability $\eta \in (0,1)$ conditionnally to $\{\rz_1,\cdots,\rz_n\}$, otherwise drawn from $P$ independently.
\end{definition}

\subsection{Membership inference security (MIS)}
\label{subsec:misdef}

We now define the MIA-wise privacy level of a learning procedure as 

\begin{definition}[MIS \citep{aubinais2023fundamental}] 
The \textit{Membership Inference Security} (MIS) of a learning procedure $\gA$ is defined as 
\begin{equation}
\label{def:sec}{\MIS}_n(P, \gA)\coloneqq c_1\left(1-c_2\;\underset{\phi}{\sup}\;{\Acc}_n(\phi; P,\gA )\right), 
\end{equation}
where $c_1$ and $c_2$ are scaling constants depending on $\lambda$ and $\eta$. The supremum is taken over all MIAs.
\end{definition}

The values of the scaling constants $c_1$ and $c_2$  are designed to ensure that the MIS ranges from $0$ to $1$. When the MIS approaches $0$, an attack that achieves almost maximum accuracy exists. Conversely, no MIA performs better than a random or constant MIA when the MIS approaches $1$.

The MIS is designed to be specific to MIAs, unlike the more general concept of differential privacy (DP)~\cite{dwork2014algorithmic}.
It can be shown that if a learning procedure is differentially private, then the MIS is lower bounded by some function of the DP parameters. However, the converse does not hold.  There are learning procedures which are not differentially private but for which the MIS approaches $1$.

One of the main results in \citep{aubinais2023fundamental} is proving that the MIS is equal to an information-theoretic quantity that measures the divergence between the joint distribution of $(\hat{\theta}_n,\rz_1)$ and the product distribution of $\hat{\theta}_n$ and an independent sample $\rz$. Here $\rz_1$ can be replaced by any $\rz_i$ of the $n$-sample, since $\hat{\theta}_n$ is a function of the empirical distribution of the $n$-sample. More details are provided in \autoref{sec:addass}.

Direct estimation of divergences between probability distributions over high-dimensional spaces is computationally prohibitive and can not be used to evaluate the  MIS. As a function of the maximum accuracy of a test function (MIA), the MIS can be evaluated as a binary classifier to get a numerical evaluation of MIS. We call it the baseline method (see~\autoref{subsec:synthetic_expe}).
However, this baseline remains computationally expensive, motivating the development of MIS evaluation techniques tailored to specific applications. In this work, we design such methods for quantized learning algorithms.

%% file: Sections/III-Theoretical_Results.tex
This section provides the main theoretical results on the MIS of quantized learning procedures. 
First, we introduce quantization procedures. Then, we present our theoretical asymptotic result, from which we deduce the definition of our new MIS indicator. Finally, we propose a method to estimate the indicator.

\subsection{Quantization}
\label{subsec:quantization}
We define a \textbf{quantizer}~\citep{citeulike:12927267} as any measurable function $\gQ:\Theta\to\Bar{\Theta}\subseteq\Theta$ for some discrete space $\Bar{\Theta}\coloneqq\{\thetaq_1,\cdots,\thetaq_K\}$.

\begin{example}[Binarized Neural Networks \citep{wang2023bitnet}]
\label{ex:BNN}
Let $\gF$ be a set of neural networks with fixed architecture.
A scalar quantizer $\gQ$ maps coordinate-wise the parameters to its sign: $\gQ(\theta) = \pp{\theta_j/|\theta_j|}_j$. Here, for any neural networks with $d$ parameters, the set $\bar{\Theta}$ would consist of all $d-$dimensional vectors in $\{-1,+1\}^d$, hence $K = 2^{d}$. 
\end{example}

\begin{example}[Vector Quantization]
Another quantization procedure, albeit underused in practice, is vector quantization. A \textit{codebook} $\Thetaq$ is usually pre-computed, which the vector quantizer $\gQ$ maps $\theta$ onto, usually performed by a nearest neighbor algorithm, which makes it efficient and memory-friendly. The constant $K$ corresponds to the number of values stored in the codebook.

\end{example}

In practice, when developing machine learning models, it is common to adjust the model's architecture based on the size of the dataset.
For example, models are often over-parameterized relative to the size of the dataset, as is the case with LLMs. 
This is why we let our quantizer \( \gQ_n  \) (and therefore the number of quantized values $K_n$ and $\Thetaq_n\coloneqq\{\thetaq^n_1,\cdots,\thetaq^n_{K_n}\}$) depend on the sample size. 
The dependence of \( \gQ_n \) on the dataset size \( n \) formalizes at least two scenarios: either the quantization procedure remains the same, but the architecture size adapts to the training dataset, or the architecture size is fixed while the quantization procedure changes. Specifically, the first interpretation can be seen as the common practice in machine learning to scale models to the dataset size.
\begin{example}[Scaling Architecture]
Let the number of parameters of our original models follow a scaling law~\citep{hoffmann2022trainingcomputeoptimallargelanguage, kaplan2020scalinglawsneurallanguage} $f$, i.e. its number of parameters is $f(n)$. Let the quantization method be fixed to a 1-bit quantization (mapping each parameter to its sign, for instance). From Example~\ref{ex:BNN}, for a dataset size $n$, the size of $\Thetaq_n$ is $K_n=2^{f(n)}$.
\end{example}

A quantizer canonically induces a \textbf{quantized learning procedure} $\gA_{\gQ_n}$. 
Additionally, we will assume without loss of generality that quantizers are numbered in ascending order
of their expected loss as follows
$$\E\cc{\ell(\thetaq^n_1,\rz)}\leq\cdots\leq \E\cc{\ell(\thetaq^n_{K_n},\rz)},
$$
where $\rz$ is a random variable with distribution $P$. 
We define the \textbf{loss gaps}
 $$\delta_k^n\coloneqq %
 \E\cc{\ell(\thetaq^n_k,\rz)}-\E\cc{\ell(\thetaq^n_1,\rz)},
 $$ 
 and the \textbf{loss variabilities}
$$\pp{\rho_{k,l}^n} = \textcolor{black}{\textrm{Cov}%
}\big(\ell\pp{\thetaq^n_k,\rz} - \ell\pp{\thetaq^n_1,\rz};\ell\pp{\thetaq^n_l,\rz} - \ell\pp{\thetaq^n_1,\rz}\big).
$$

\subsection{Asymptotic theorem and the new MIS indicator}

The following asymptotic result is a consequence of the moderate deviation principle. 
One of the main difficulties in proving this theorem was establishing the appropriate framework and assumptions to apply the principle. We now present the assumptions.\\

\textbf{(A1)} We have ${\mkn{2}\toinf{n}0}$ and ${\sqrt{n}\mkn{2}\toinf{n}+\infty}$.\\
\textbf{(A2)}  
  a) The limits $\rho_{k,l} = \liminf{n}\rho_{k,l}^n$ and
    $c_k = \liminf{n} \frac{\delta_2^n}{\delta_k^n}$ exist. \\
    b) There exists $K<\infty$ such that for all $k>K$, $c_k^2\sigma_{k}^2=0$. Here, $\sigma_{k}^2=\rho_{k,k}$.\\
    c) The matrix $\Lambda=(\rho_{k,l}c_{k}c_{l})_{2\leq k,l \leq K}$ is non singular.\\
\textbf{(A3)} For all $t>0$, there exists $M$ such that for all $n$,  %
$\E\cc{\exp t\left\|\pp{\frac{\delta_2^n}{\delta_k^n}\big(\ell\pp{\thetaq^n_k,\rz} - \ell\pp{\thetaq^n_1,\rz}\big)} _{1<k\leq K_n}\right\|_2}\leq M$.\\
\textbf{(A4)} The sequence $\pp{\pp{\frac{\delta_2^n}{\delta_k^n}\big(\ell\pp{\thetaq^n_k,\rz} - \ell\pp{\thetaq^n_1,\rz}}_{1<k\leq K_n}}_n$,  in $l_2(\R)$, of random variables, is tight.
\\

Assumptions \textbf{(A1)} and \textbf{(A2)} depend on the quantization procedure.
They indicate that the quantization procedure has two important properties: the expected loss landscape near its minimum can be well described by multiple quantized parameter configurations; the per-sample losses have enough variability across the quantized parameters. Assumption \textbf{(A2)} b) is empirically supported by our experiments (see the supplementary materials~\autoref{app:molecular_details} in~\autoref{fig:k_max}).
When the loss function is bounded, \textbf{(A3)} is satisfied, and \textbf{(A4)} and  \textbf{(A2)} a) are light assumptions.

\begin{theorem}\label{thm:seqQ} %
Under the assumptions \textbf{(A1), (A2), (A3), (A4)} we have
\begin{multline*}
    \limsup_{n\to\infty}{\frac{1}{n\pp{\mkn{2}}^2}\log\big(1-\MIS_n(P, \gA_{\gQ_n})\big)}\\
    \leq  - \frac{1}{2\; \underset{k}{\max} \;c_k^2  \sigma_{k}^2} < 0.
\end{multline*}
\end{theorem}
The fact that  $\underset{k}{\max} \;c_k^2  \sigma_k^2$ is finite follows from \textbf{(A2)}.  \autoref{thm:seqQ} shows that the MIS of the quantization procedure approaches $1$ when the sample size $n\rightarrow \infty$, thanks to \textbf{(A1)},  at a specified rate, controlled by
$$\bar{r}_{\mathcal{Q}_n} \coloneqq \pp{\delta_2^n}^2/(2\;\underset{k}{\max} \;c_k^2  \sigma_{k}^2).$$
\autoref{thm:seqQ} suggests that for quantization procedures $\gQ_n$ and $\gR_n$,  if $\bar{r}_{\mathcal{Q}_n} \geq \bar{r}_{\mathcal{R}_n}$, then $\gA_{\gQ_n}$ produces more secure parameters than $\gA_{\gR_n}$ for a large enough sample size.
We use this rate to define the new MIS indicator, in which the limits are approximated by the value at the sample size $n$.
\begin{definition}[MIS indicator]
\label{Indicator}
The MIS indicator of the quantization procedure $\gQ_n$ is defined as
$$\qcertif\coloneqq \frac{1}{2}\underset{k>1}{\min}\frac{\cc{\E\cc{\ell(\thetaq^n_k,\rz)}-\E\cc{\ell(\thetaq^n_{1},\rz)}}^2}{\textrm{Var}\cc{\ell(\thetaq^n_k,\rz)-\ell(\thetaq^n_{1},\rz)}}.$$ 
\end{definition}

Note that the MIS indicator wholly relies upon the random variables $\ell(\thetaq_k^n,\rz) - \ell(\thetaq_1^n,\rz)$, making it an attack-free privacy indicator. The next step is to be able to compute the indicator.

%% file: Sections/IV-Algorithm.tex
\label{ssec:algo}

To estimate the MIS indicator of a quantization procedure $\gQ_n$
we must compute the loss gaps $\mkn{k}$ for all $k\leq K_n$.
However, this is computationally infeasible, as even a simple quantizer like 1-bit quantization leads to an exponentially large $K_n$. %
We observe that $\max_k\pp{\mkn{2}/\mkn{k}}^2(\sigma_k^n)^2$ is empirically dominated by low-loss quantized models (see~\autoref{subsec:eval_privacy}). This suggests that exploring the entire set $\Bar{\Theta}$ is unnecessary; focusing on low-loss quantized models is sufficient to estimate $r_{\gQ_n}$.

We thus propose estimating using quantized models derived from $\hat{\theta}_n$'s training trajectory.
Since the quantizers with the lowest loss likely reside near the trajectory of the minimized loss, we limit our computations to these weights to efficiently capture critical quantizers.
The estimation procedure given the list of quantized losses is outlined in \ccref{Algorithm}{algo:main}, and additional details are given in~\autoref{app:algo}.
Expectations and variances are estimated by the empirical estimators using the sample set. Finally, to obtain more accurate estimations of $\qcertif$, we average the estimated value over multiple training trajectories (we analyzed the influence of the number of trajectories in \autoref{subsec:synthetic_expe}).

\begin{algorithm}
    \caption{Estimation of $\qcertif$ from the quantized losses}
    \label{algo:main}
    \begin{algorithmic}[1]
        \STATE \textbf{Input:} The set of each sample's loss for the $K$ checkpoints: $\Ls \in \mathbb{R}^{K\times n}$, and the average loss of each checkpoint $m\in\mathbb{R}^{K}$
        \STATE \textbf{Output:} An estimate of $\qcertif$.
        \STATE $\text{idx} \leftarrow \text{argsort}(m)$. // Sort the quantized models
        \STATE $m\leftarrow m[\text{idx}]$.
        \STATE $\Ls \leftarrow \Ls[\text{idx}]$.
        \FOR{$k=2$ to K}
            \STATE $\sigma^2[k] \leftarrow \textrm{Var}\pp{\Ls[k] - \Ls[1]}$.
        \ENDFOR
        \STATE \textbf{return} $r_{\gQ} \leftarrow \frac{1}{2}\cc{\underset{2\leq k\leq K}\max{\frac{\sigma^2[k] }{\pp{m[k] - m[1]}^2}}}^{-1}$.

    \end{algorithmic}
\end{algorithm}

%% file: Sections/V-Numerical_Experiments.tex
\section{NUMERICAL EXPERIMENTS}
\label{sec:num_exp}

\begin{figure}
    \centering
    \includegraphics[width=1\linewidth]{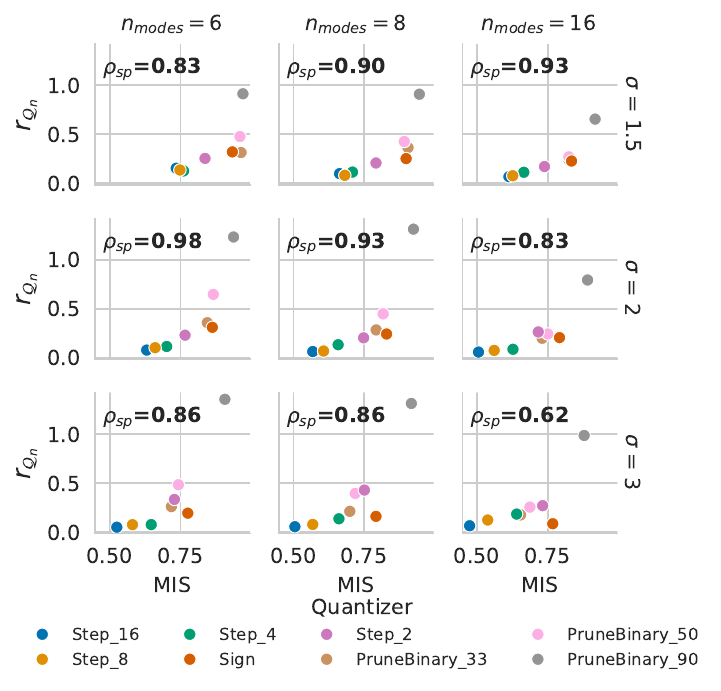}
    \caption{
        \textbf{Relationship between $\qcertif$ and the MIS.}
         Estimated values of $\qcertif$ and the MIS for each quantizers on different datasets, with their corresponding Spearman correlation ($\rho_{sp}$).
    }
    \label{fig:scatterplot_synth}
\end{figure}

\subsection{Validation of \texorpdfstring{$\qcertif$}{qcertif} as a MIS indicator}
\label{subsec:synthetic_expe}

To validate that our estimation of $\qcertif$ effectively ranks quantization methods by their privacy level, we compare the resulting ordering with the baseline MIS-based ranking obtained from synthetic experiments.
As we focus solely on ranking quantization procedures, we omit specific 
$\qcertif$ values in the figures for clarity and readability.

\paragraph{MIS Baseline.}
To evaluate the ability of our estimator to correctly rank quantizers in terms of membership security, we rely on a baseline estimation of the MIS (theoretical details given in \autoref{app:baseline_estimation}).
This baseline consists of training a discriminator $\bsl$ to differentiate between samples from the training set and other samples given the model's weights.

To train the classifier, we follow the following steps:
\begin{enumerate}
    \item We draw $n=128$ samples $(x,y)$ from our synthetic dataset, and train a classifier $\thetan$ on these samples.
    \item We draw $n$ samples, that are the samples out of $\thetan$'s training set.
    \item We create the sub-dataset built with $\thetan$:  \(
        \{
            (\thetan, (x, y)_{i}, t_{i})
        \}_{
           i\leq 2n
        }
    \) where $t_{i}$ equals $1$ if the sample $(x,y)_{i}$ is in $\thetan$'s training set, and $0$ otherwise.
    \item We perform these operations k=$300$ times and create the full dataset as the concatenation of all sets described in step 2, which we split into a training and testing set, to train and evaluate $\bsl$.
\end{enumerate}

By construction, all samples $(x,y)$ are drawn independently, and in particular all samples used to evaluate $\bsl$ are independent from all samples in its training set. This property is particularly data intensive and is the reason why we focused on synthetic data in this section.
Finally, the baseline approach measures the MIS as
$MIS_{\bsl} = 2-2\text{Acc} \pp{\bsl, P, \gA}$ (more details are provided in~\autoref{app:baseline_estimation}).

\paragraph{Datasets and Classifiers.}
We generate data points sampled from $\mathbb{R}^{128}$ using $k_{\textrm{modes}} \in \{6,8,16\}$ isotropic Gaussian distributions of standard deviation $\sigma\in \{1.5,2,3\}$, where each cluster is assigned a label in $\{0,1\}$.
We trained each classifier (single-layer fully connected networks) on $n=128$ samples using the Adam optimizer with a learning rate of $10^{-4}$, and we stopped the training at 30 epochs (see the training curves in~\autoref{app:synthetic_details}).

\begin{figure}
    \centering
    \includegraphics[
        width=1.\linewidth,]{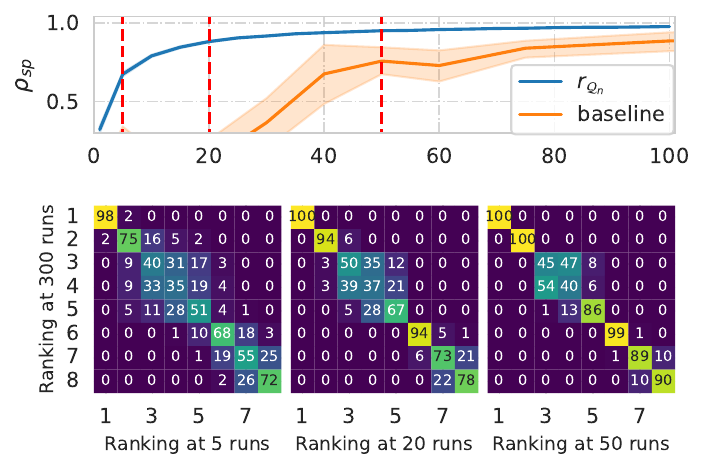}
    \caption{
        \textbf{Stability of $\qcertif$'s estimation.}
        Spearman correlation between rankings obtained with a limited number of runs and those obtained with the full set of 300 runs. The ranking induced by $\qcertif$ stabilizes after about 15 runs, while the baseline MIS requires substantially more runs to converge to the final ranking.
    }
    \label{fig:sta}
\end{figure}

\paragraph{Quantization.}
For these experiments, we consider a range of different simple quantization methods, including: 1bit quantization by taking the sign of the weights (\texttt{Sign}), 1.58 bits quantization with different sparsity levels (\texttt{1.58b \{x\}\%} where $x\%$ of the weights with the smallest magnitude are set to zero, and the rest to their sign), and quantization from 2 to 5 bits.

\begin{figure*}
    \centering
    \includegraphics[
        width=0.95\linewidth, trim=0.25cm 0.25cm 0.3cm 0.1cm, clip,
    ]{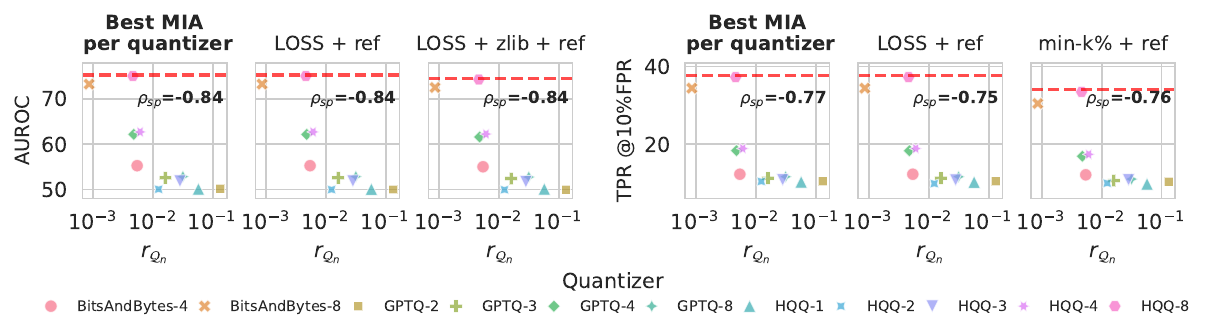}
    \caption{
        \textbf{Quantized security against the performances of various black-box MIAs for language models.}
        The first plot shows the relation between $\qcertif$ the performances of the best MIA for each quantizer.
        As $\qcertif$ increases, the performances of the MIAs decreases, indicating that the quantized model is more secure.
        The dashed red line indicates the performance of the different MIAs on the full-precision model.
    }
    \label{fig:NLP_res}
\end{figure*}

\paragraph{Estimation of the privacy guarantees.}
\autoref{fig:scatterplot_synth} illustrates the correlation between our proposed metric $\qcertif$ and the MIS baseline.
We report the Spearman correlation between both metrics (the correlation between the rankings induced by both metrics), and as expected, quantizers with more bits of information, such as \texttt{5 bits} and \texttt{4 bits}, are the least private.
In contrast, the \texttt{1.58b 90\%} quantizer, which introduces 90\% sparsity by setting weights to zero, achieves the highest privacy.
Overall, the rankings produced by $\qcertif$ closely match those of the baseline method, with an average Spearman correlation of $\rho_{sp}=0.86$, demonstrating that $\qcertif$ reliably ranks quantization methods by their privacy levels.

\paragraph{Stability of the rankings.}
In~\autoref{fig:sta}, we examine how the number of runs influences the stability of the ranking by comparing the rankings obtained with subsets of runs ($k_{\text{runs}} \leq 300$) against those obtained using all 300 runs. 
For both $\qcertif$ and the baseline MIS evaluation, increasing the number of runs ($\thetan$) yields rankings that better align with the rankings using all runs.
Notably, $\qcertif$ achieves stability much faster: after only 15 runs, its rankings already exhibit a high correlation with the 300-run ranking ($\rho_{sp} \geq 0.9$), whereas the baseline MIS requires around 150 runs to reach the same correlation level.

\subsection{Validation on practical use-case: Language models for drug discovery}%
\label{subsec:text_expe}

As a second experimental setting, we consider a real-world application in drug discovery.
In this domain, data is both highly valuable and sensitive, making it crucial to assess whether predictive models might inadvertently leak proprietary information.
We evaluate the privacy of generative language models trained on chemical knowledge, comparing our proposed measure $\qcertif$ against standard membership inference attacks (MIAs) for language models.

\paragraph{Experimental setup.}
We fine-tuned 20 models based on Qwen2.5-instruct-0.5B~\citep{qwen} using LORA adapters of rank 64~\citep{hu2021loralowrankadaptationlarge} on the SMolInstruct dataset~\citep{smol} ($\approx$10k samples per model), for 10 epochs.
We considered the molecular captioning and generation tasks, consisting of either giving a description of a given molecule, or generating a molecule from a given description. Results on Ministral-7B and Llama3-8B are given in \autoref{fig:big_llms}.

\paragraph{Quantizers.}
We evaluate the value of $\qcertif$ for various classical quantization methods including BitsAndBytes~\citep{bab} (4-8 bits), GPTQ~\citep{gptq} (2-3-4-8 bits), and HQQ~\citep{hqq} (1-2-3-4-8 bits).
We evaluate the per-sample loss of the quantized model on 22 checkpoints of the model to estimate $\qcertif$.

\paragraph{Black-box results.}
We compare the evolution of our privacy metric $\qcertif$ with the performance of several black-box MIAs on the quantized models (see~\autoref{app:NLP} for details on the selected attacks).
\autoref{fig:NLP_res} shows the relationship between $\qcertif$ and MIA performance, including the best-performing attack on each quantizer.
We find that higher values of $\qcertif$ correspond to lower MIA performance, indicating stronger privacy guarantees.
Conversely, lower values of $\qcertif$ are associated with higher MIA performance.
Overall, these results demonstrate that $\qcertif$ successfully orders quantization methods according to their privacy level.

\begin{figure}
    \centering
    \includegraphics[
        width=0.95\linewidth, trim=0.25cm 0.25cm 0.3cm 0.1cm, clip,
    ]{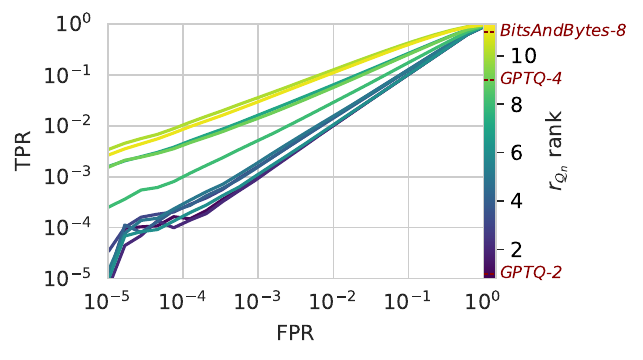}
    \caption{
        \textbf{TPR-FPR curves of LiRa attacks.}
        True positive rate (TPR) of a LiRa offline attack at different low false positive rate (FPR).
        Models that are hardest to attack (resp. easiest to attack) are also those ranked as more (resp. least) secure by $\qcertif$.
    }
    \label{fig:lira}
\end{figure}

\paragraph{LiRa attack.}
We further compare the privacy rankings of the quantizers with the performance of one of the strongest MIAs: the Likelihood Ratio attack (LiRa)~\cite{carlini2022membership,hayes2025strongmembershipinferenceattacks}.
This attack requires training a set of reference models on random subsets of the data, and then comparing their outputs on a sample with that of the target model (model under attack).
In its online form, the target model is compared to $N$ reference models trained without the sample of interest and to $N$ reference models trained with it.
Because of its computational cost, an offline variant has been proposed, which relies only on models trained without the sample of interest.
We adopt this offline setting with $N=35$, which we found sufficient to obtain effective attacks.
\autoref{fig:lira} reports the true positive rate (TPR) at varying false positive rates (FPR) for all quantizers, and shows that the quantizers hardest to attack are also those ranked as most secure by $\qcertif$, further validating our metric.

\subsection{Security performance tradeoff on molecular property prediction}
\label{subsec:molecular_expe}

\begin{figure}
   \centering
   \includegraphics[
        width=0.95\linewidth, trim=0.25cm 0.25cm 0.2cm 0.cm, clip,
    ]{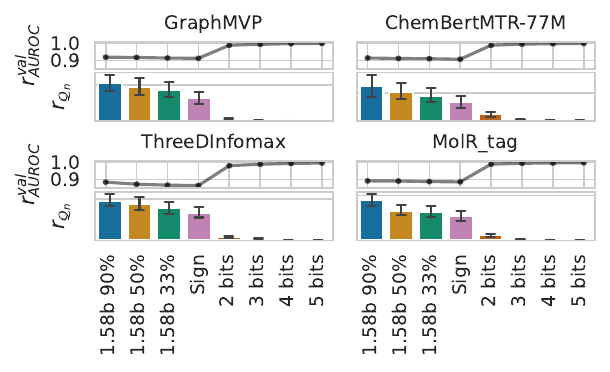}
   \caption{
       \textbf{Impact of quantization on classification tasks.}
       Evolution of each downstream model's privacy $\qcertif$ along with relative performances of the quantized models compared to the original on classification tasks.
   }
   \label{fig:barplot_dperfs_vs_rdelta}
\end{figure}

Finally, we study the security of small predictive models trained on molecular embedders, focusing on the trade-off between performance and privacy.

\paragraph{Pretrained Embedders.}
To generate one-dimensional molecular embeddings, we evaluate four pretrained models from the representation learning literature: GraphMVP~\citep{liu2022pretraining} and 3D-Infomax~\citep{stark2021_3dinfomax} (3D-2D mutual information maximization), MolR~\citep{wang2022chemicalreactionaware} (reaction-aware pretraining), and ChemBERTa-MTR~\citep{ahmad2022chemberta2} (multitask regression with string representations of the molecule).
The embeddings are passed to two layers feed-forward networks (with a hidden dimension of 128) trained on each downstream task.
We trained all models for 500 epochs with a learning rate of 1e-4, and selected the best checkpoint based on the validation loss.
We focused on the simple quantization methods presented in~\autoref{app:quantization} for their interpretability.

\paragraph{Downstream tasks.}
We evaluate and train the models on various property prediction tasks (classification) from the Therapeutic Data Commons (TDC) platform~\citep{Huang2021tdc}, focusing on ADMET properties (Absorption, Distribution, Metabolism, Excretion, and Toxicity).
Additional results and tradeoffs with regression tasks are given in~\autoref{app:molecular_details}.
We report the AUC-ROC scores and to ensure robust estimation of $\qcertif$, we allocate 40\% of the dataset to the validation set.
All models are trained using 90\% of the dataset for each run to ensure different training trajectories are sampled.
Finally, we report the relative performance of the quantized model to the original: \(
\diffmetric{\textrm{val}}{\text{AUROC}} = {\text{AUROC}^{\textrm{val}}_{\text{quantized}}}/{\text{AUROC}^{\textrm{val}}_{\text{original}}}
\).

\paragraph{Performance vs Privacy trade-off.}
\autoref{fig:barplot_dperfs_vs_rdelta} shows the trade-off between privacy ($\qcertif$) and average task performance across embedders.
Even under the strongest quantization, AUROC drops by only about 10\%, likely because classification primarily depends on boundary regions that seem not to be too affected by  quantization procedures.
We see that higher downstream accuracy generally comes at the expense of lower privacy.
A threshold effect emerges: quantization at 2 bits or more preserves performance but does not improve privacy, whereas stronger quantization (1.58 bits) increases security at the cost of roughly 10\% AUROC.

%% file: Sections/VI-Limitations.tex
In this work, we investigated Membership Inference Security of quantization procedures for machine learning models that minimize an empirical loss. We proved that the learning procedure is asymptotically determined by the distribution of the quantized models’ loss per sample. From this, we derived a new MIS indicator.
We introduced an attack-free methodology for comparing quantization procedures in terms of MIS, with limited computational cost. Through extensive experiments on both synthetic and real-world datasets, we validated the effectiveness of our approach.

We believe our work suggests several interesting new research directions related to the limitations of our paper.
First, we defined MIS as quantifying the security against any possible MIA. In some cases, however, not all MIAs are threats. 
The supremum in~\autoref{def:sec} could be taken only over a subset of all MIAs; for example, consider black-box attacks.
Additionally, our study focused on post-training quantized models and did not identify any quantizer  that can achieve strong downstream performance and high security simultaneously. One interesting direction for future research is to propose Quantization Aware Training methods to develop quantizers that are trained to maximize $\qcertif$ while being jointly optimized with the task-specific loss function. This is an area of ongoing research.

%% file: Appendix/XVI-Table_of_Notations.tex
\clearpage
\section{Notations}
\label{sec:notations}

\renewcommand{\arraystretch}{1.2}

\begin{table}[h!]
\centering
\resizebox{\textwidth}{!}{
    \begin{tabular}{|c|l|c|c|}
    \hline
    \textbf{Variable} & \textbf{Description} & \textbf{Mathematical Definition} & \textbf{First introduction} \\
    \hline 
    
    $P$ & data distribution & -- & \multirow{11}{*}{\autoref{subsec:notations}}\\

    \cline{1-3}
    $\gZ$ & data space & -- & \\
    
    \cline{1-3}
    $\rz_j$ & training sample & $\rz_j\sim P$ & \\ 
    
    \cline{1-3}
    
    $n$ & training dataset size & -- &  \\ 
    
    \cline{1-3} 
    
    $\gD_n$ & training dataset & -- &  \\ 
    
    \cline{1-3}
    
    $\hat{P}_n$ & empirical distribution & $\frac{1}{n}\sum_{j=1}^{n}\delta_{\rz_j}$ & \\
    
    \cline{1-3}
    
    $\Psi_\theta$ & predictor with parameters $\theta$ & -- &  \\ 
    
    \cline{1-3}
    
    $\Theta$ or $\Theta_n$ & set of parameters & $\Theta\subseteq\mathbb{R}^d$ &  \\
    
    \cline{1-3} 
    
    $\gF$ & set of predictors & $\{\Psi_\theta : \theta\in\Theta\}$ & \\ 
    
    \cline{1-3} 
    
    $\ell$ & loss function & $\ell:\Theta\times\gZ\to\mathbb{R}^+$ &  \\ 
    
    \cline{1-3} 
    
    $\gA$ & learning procedure & $\gA:\bigcup_{n\geq1}\gZ^n\to\Theta$ & \\ 
    
    \hline 
    
    $\phi$ & MIA & $\phi:\Theta\times\gZ\to\{0, 1\}$ & \multirow{5}{*}{\autoref{subsec:miaacc}} \\ 
    
    \cline{1-3}
    
    $\tilde\rz$ & test sample & -- & \\ 
    
    \cline{1-3}
    
    $\eta$ & probability of $\tilde\rz$ being member & $\mathbb{P}(\tilde\rz\text{ is member})$ & \\ 
    
    \cline{1-3}
    
    $\lambda$ & importance of the TPR & -- & \\ 
    
    \cline{1-3}
    
    $\Acc_n(\phi;P,\gA)$ & accuracy of an MIA & $\mathbb{E}[\texttt{TNR}] + \lambda\mathbb{E}[\texttt{TPR}]$ & \\ 
    
    \hline
    
    $\MIS_n(P,\gA)$ & Membership Inference Security & $c_1(1-c_2\sup_\phi\;\Acc_n(\phi;P,\gA))$ &  \multirow{1}{*}{\autoref{subsec:misdef}}\\ 
    
    \hline
    $\bar{\Theta}_n$ & discrete set of parameters & $\bar{\Theta}_n=\{\bar\theta_1^n,\ldots,\bar\theta_{K_n}^n\}\subseteq\Theta$ & \multirow{11}{*}{\autoref{sec:theo_res}} \\ 
    
    \cline{1-3}
    
    $\gQ_n$ & quantizer & $\gQ_n:\Theta\to\bar{\Theta}_n$ &  \\

    \cline{1-3} 
    
    $\mathbb{P}_\rx, \mathbb{P}_{(\rx_1,\rx_2)}, \mathbb{P}_{\rx_1}\otimes\mathbb{P}_{\rx_2}$ & marginal, joint, product distribution & -- &  \\ 
    
    \cline{1-3}

    $\delta_k^n$ & loss gap & $\mathbb{E}[\ell(\bar\theta_k^n,\rz)]-\mathbb{E}[\ell(\bar\theta_1^n,\rz)]$ &  \\ 
    
    \cline{1-3} 
    
    $\rho_{k,l}^n$ & $(k,l)$-loss variability & $Cov(\ell(\bar\theta_k^n,\rz) - \ell(\bar\theta_1^n,\rz),\ell(\bar\theta_l^n,\rz) - \ell(\bar\theta_1^n,\rz))$ &  \\ 
    
    \cline{1-3}

    $\rho_{k,l}$ & asymptotic $(k,l)$-loss variability& $\underset{n\to\infty}{\lim}\rho_{k,l}^n$ & \\ 

    \cline{1-3}

    $\sigma_k^2$ & asymptotic $(k,k)$-loss variability & $\rho_{k,k}$ & \\

    \cline{1-3}

    $c_k$ & asymptotic ratio between gaps & $\underset{n\to\infty}{\lim} \delta_2^n/\delta_k^n$ & \\ 

    \cline{1-3} 

    $\Lambda$ & matrix  & $(\rho_{k,l}c_kc_l)_{2\leq k,l\leq K}$ & \\ 
    \cline{1-3}

    $\bar{r}_{\gQ}^{n}$ & log rate of \autoref{thm:seqQ} & $(\delta_2^n)^2 / (2 \max_k c_k^2 \sigma_k^2)$ & \\ 

    \cline{1-3}

    $r_{\gQ_n}$ & MIS indicator & $\frac{1}{2}\underset{k>1}{\min}\frac{\cc{\E\cc{\ell(\thetaq^n_k,\rz)}-\E\cc{\ell(\thetaq^n_{1},\rz)}}^2}{\textrm{Var}\cc{\ell(\thetaq^n_k,\rz)-\ell(\thetaq^n_{1},\rz)}}$ & \\
    
    \bottomrule

    \end{tabular}
}
\vspace{0.5em} 
\caption{Table of Notations}
\end{table}
\newpage

%% file: Appendix/XI-A.tex
\section{MIS is related to an f-divergence}
\label{sec:addass}

We recall here the main theoretical result of \cite{aubinais2023fundamental}. %

Let $\phi$ be an MIA, and consider its accuracy ${\text{Acc}_n}(\phi; P,\gA)$. 
Let $\hat{P}_n$ be the empirical distribution of the random variables $\rz_1,\ldots,\rz_n$.
We define a test point $\Tilde{\rz}$ as follows. Let $T\in\{0,1\}$ be a Bernoulli random variable with parameter $\eta\in(0,1)$. Let $\rz_0\sim P$ and $U$ be a random variable whose distribution is $\hat{P}_n$ conditionally to 
$\rz_1,\ldots,\rz_n$, where $T, \rz_0$ and $U$ are independent. Additionally, $\rz_0$ and $T$ are independent of %
$\rz_1,\ldots,\rz_n$. We define the test sample by,

\begin{equation}
\label{eq:test_pt}
\Tilde{\rz}\coloneqq TU + (1-T)\rz_0.
\end{equation}

This definition is a formalization of the fact that a test sample is either a member of the training dataset ($T=1$) with probability  $\mathbb{P}(T=1) = \mathbb{P}(\tilde\rz\text{ is member})=\eta$, or not a member ($T=0$). The random variable $\rz_0$ is used as a placeholder to represent the non-member property of the test point and $\eta$ represents the theoretical fraction of members. Using \autoref{eq:test_pt}, the accuracy of $\phi$ can be rewritten as 

\begin{align}
    {\text{Acc}_n}(\phi; P,\gA) &= \mathbb{P}(\phi(\hat{\theta}_n, \rz_0)=0, T=0) + \lambda \mathbb{P}(\phi(\hat{\theta}_n, \rz_1)=1, T=1) \\ 
    \nonumber&= (1-\eta)\mathbb{P}(\phi(\hat{\theta}_n,\rz_0)=0) + \lambda\eta\mathbb{P}(\phi(\hat{\theta}_n,\rz_1)=1).
\end{align}

Interestingly, constant MIAs $\phi_0 \equiv 0$ and $\phi_1\equiv1$ have respectively an accuracy of $1-\eta$ and $\lambda\eta$, which means that any MIA whose accuracy is worse than $\max(1-\eta,\lambda\eta)$ is irrelevant to use. Additionally, if a perfect MIA $\phi^*$ exists, then its accuracy would equal $1-\eta + \lambda\eta$. This means that the supremum over all MIAs of the accuracy is bounded as 
\begin{equation}
    \max(1-\eta,\lambda\eta)\leq \sup_\phi {\text{Acc}_n}(\phi; P,\gA) \leq 1-\eta + \lambda\eta.
\end{equation}

From this equation, the MIS can be defined as follows,

\begin{definition}[MIS \citep{aubinais2023fundamental}] The membership inference security of a learning procedure $\gA$ is 

\begin{equation}
    {\MIS}_n(P, \gA) \coloneqq \frac{1}{\min(1-\eta, \lambda\eta)}\left(1-\eta + \lambda\eta - \sup_\phi {\text{Acc}_n}(\phi; P,\gA)\right),
\end{equation}

where the supremum is taken over all MIAs.    
\end{definition}

Thus,  $c_1 = (1-\eta + \lambda\eta)/\min(1-\eta,\lambda\eta)$ and $c_2 = (1-\eta+\lambda\eta)^{-1}$ in \eqref{def:sec}.
\\

For any $\alpha>0$, and any distributions $P$ and $Q$ over the same probability space, we define %
\begin{align*}
    \tilde D_\alpha(P,Q) &\coloneqq \frac{1}{\alpha}\underset{B}{\sup}\left[\alpha P(B) - Q(B)\right]\\
    D_\alpha(P,Q)&\coloneqq \max(1,\alpha)\left[\tilde D_\alpha(P,Q) - \left(1-\frac{1}{\alpha}\right)_+\right],
\end{align*}
where for any real number $x\in\mathbb{R}$, we have $x_+ = \max(0,x)$.  Here, the supremum is taken over all measurable sets. 

One of the main results in \cite{aubinais2023fundamental} is to prove that the MIS is related to an f-divergence between the joint distribution of $(\hat{\theta}_n,\rz_1)$ and the product distribution of $\hat{\theta}_n$ and an independent sample $\rz_0$. The setting in \cite{aubinais2023fundamental} is more general and \autoref{thm:aub} below holds for any possibly randomized algorithm as soon as it is a (possibly randomized) function of $\hat{P}_n$. 
\begin{theorem}[Theorem 6 and Proposition 5 of \cite{aubinais2023fundamental}]
\label{thm:aub}
Let $\gamma\coloneqq \frac{1-\eta}{\lambda\eta}$. For any distribution $P$ and any learning procedure $\gA$, we have

\begin{equation}
\label{eq:mis_delta}
{\MIS}_n(P, \gA) = 1 - D_\gamma\left(\mathbb{P}_{(\hat{\theta}_n,\rz_0)},\mathbb{P}_{(\hat{\theta}_n,\rz_1)}\right),
\end{equation}
where for any random variable $\rx$, $\mathbb{P}_\rx$ designates its distribution. \\
Additionally, the map $(P,Q)\mapsto D_\gamma(P,Q)$ is an $f-$divergence and for any random variables $\rx_1$ and $\rx_2$ with joint distribution $\mathbb{P}_{(\rx_1,\rx_2)}$, it holds that 

\begin{equation}
    D_\gamma(\mathbb{P}_{\rx_1}\otimes\mathbb{P}_{\rx_2}, \mathbb{P}_{(\rx_1,\rx_2)}) = \mathbb{E}_{\rx_2}\left[D_\gamma(\mathbb{P}_{\rx_1}, \mathbb{P}_{\rx_1\mid\rx_2})\right],
\end{equation}

where $\mathbb{P}_{\rx_1\mid\rx_2}$ is the distribution of $\rx_1$ conditionally to $\rx_2$.
\end{theorem}
Notice that in \autoref{eq:mis_delta}, $\rz_1$ could be replaced by any $\rz_j$, for $j\in\{1,\ldots,n\}$ since $\hat{\theta}_n$ is a function of $\hat{P}_n$.

We set
\begin{equation}
\label{eq:def_deltagen}
    \Deltagen(P,\gA) \coloneqq D_\gamma\left(\mathbb{P}_{(\hat{\theta}_n,\rz_0)},\mathbb{P}_{(\hat{\theta}_n,\rz_1)}\right).
\end{equation}

Additionally, setting $p_{(\hat{\theta}_n,\rz_0)}$ (resp. $p_{(\hat{\theta}_n,\rz_1)}$) the density of $\mathbb{P}_{(\hat{\theta}_n,\rz_0)}$ (resp. $\mathbb{P}_{(\hat{\theta}_n,\rz_1)}$) with respect to a dominating measure $\zeta$, $\Delta_{\lambda,\eta,n}(P,\gA)$ can be written in integral form as 

\begin{align}
  \label{eq:integral_form}  \Deltagen(P,\gA) &= \max(1,\gamma)\left[\frac{1}{2\gamma}\int|\gamma p_{(\hat{\theta}_n,\rz_0)} - p_{(\hat{\theta}_n,\rz_1)}|d\zeta - \frac{1}{2}\left|1-\frac{1}{\gamma}\right|\right] \\ 
   \label{eq:integral_form_expectation} &= \mathbb{E}_{\rz_1}\left[\max(1,\gamma)\left[\frac{1}{2\gamma}\int|\gamma p_{\hat{\theta}_n} - p_{\hat{\theta}_n \mid\rz_1}|d\zeta - \frac{1}{2}\left|1-\frac{1}{\gamma}\right|\right]\right],
\end{align}

where $p_{\hat{\theta}_n}$ (resp. $p_{\hat{\theta}_n\mid\rz_1}$) is the density of the marginal distribution (resp. conditional distribution given $\rz_1$) of $\hat{\theta}_n$.

\section{Proof of \autoref{thm:seqQ}}

We introduce  additional notations. For $i=1,\ldots,K_n$, we define
$m_{\thetaq_{i}^n} =\E\pp{\ell(\thetaq_k^n,\rz_i)} $, so that $m_{\thetaq_{1}^n}<\cdots<m_{\thetaq_{K_n}^n}$. For $i=1,\ldots,K_n$ and $j=1,\ldots,n$, we define $L_{k,j}^n\coloneqq \ell(\thetaq_k^n,\rz_j) - \ell(\thetaq_1^n,\rz_j)$. We recall that the \textbf{loss gaps} are defined as $\delta_k^n = \mathbb{E}[L_{k,1}^n]=m_{\thetaq_{k}^n}-m_{\thetaq_{1}^n}$
and the \textbf{loss variabilities} as $\pp{\sigma_k^n}^2\coloneqq \textrm{Var}\pp{L_{k,1}^n}$. Finally we set $D^n\coloneqq \textrm{diag}\pp{\pp{\delta_k^n}_{k>1}}$, the $K_{n}\times K_{n}$-diagonal matrix with diagonal entries the $\delta_k^n$, $k=1,\ldots,K_n$.

We first prove that the divergence $D_{\alpha}$ is dominated by the total variation distance.

\begin{proposition}
    \label{prop:TV}
    For any $\alpha>0$, and any distributions $P$ and $Q$ over the same probability space,
    $$D_{\alpha}(P,Q) \leq D_{1}(P,Q)=\|P-Q\|_{TV}
    $$
 where   $\|P-Q\|_{TV}$ is the total variation distance between $P$ and $Q$.
\end{proposition}

\begin{proof}[Proof of Proposition \ref{prop:TV}]
Usual manipulations allow to prove that, if $p$ (resp. $q$) is the density of $P$ with respect to a dominating measure $\zeta$, then
$$
D_{\alpha}(P,Q)=\left\{\begin{array}{ll}
\int \left(p-\frac{1}{\alpha}q\right)_{+}d\zeta &\;\text{if}\;\alpha \leq 1\\
\int \left(q- \alpha p\right)_{+}d\zeta  &\;\text{if}\;\alpha \geq 1.
\end{array}
\right.
$$
But $\alpha \mapsto \left(p-\frac{1}{\alpha}q\right)$ is non decreasing, so that the sets $\left(p-\frac{1}{\alpha}q \geq 0\right)$ are non decreasing also, so that for all $\alpha \leq 1$, 
$\int \left(p-\frac{1}{\alpha}q\right)_{+}d\zeta \leq \int \left(p-q\right)_{+}d\zeta $. In the same way,
$\alpha \mapsto \left(q- \alpha p\right)$ is non increasing, so that the sets $\left(q- \alpha p \geq 0\right)$ are non increasing also, so that for all $\alpha \geq 1$, 
$\int \left(q- \alpha p\right)_{+}d\zeta \leq \int \left(p-q\right)_{+}d\zeta $. The conclusion follows.

\end{proof}
Recall $\gQ_n$ is a  quantizer and
$\hat{\theta}_n=\gA_{\gQ_n}\pp{\rz_1,\cdots,\rz_n}$.
Applying \autoref{thm:aub} and Proposition \ref{prop:TV}, we get
\begin{equation}
\label{ineq:mis_delta}
1 - {\MIS}_n(P, \gA) \leq \|\mathbb{P}_{(\hat{\theta}_n,\rz_0)}-\mathbb{P}_{(\hat{\theta}_n,\rz_1)}\|_{TV},
\end{equation}
and \autoref{thm:seqQ} follows from Proposition \ref{prop:1}
and Proposition \ref{prop:2} below.
\begin{proposition}
\label{prop:1}
Under the assumptions of of \autoref{thm:seqQ}, we have 

\begin{equation}
    \limsup_{n\to\infty}\frac{1}{n \pp{\mkn{2}}^2}\log \|\mathbb{P}_{(\hat{\theta}_n,\rz_0)}-\mathbb{P}_{(\hat{\theta}_n,\rz_1)}\|_{TV} \leq -\underset{x\in\Omega^c}{\inf}\frac{x^T\Lambda^{-1}x}{2},    
\end{equation}
where $\Lambda$ is defined as
\begin{align*}
     \Lambda &\coloneqq \begin{pmatrix}
        \rho_{2,2}c_2^2 & \cdots & \rho_{2, K}c_2c_K \\ 
        \vdots & \ddots & \vdots \\ 
       \rho_{K, 2}c_{K}c_2 & \cdots & \rho_{K,K}c_K^2
    \end{pmatrix},
\end{align*}
with %
$\Omega \coloneqq [-1,\infty)^{K-1}$.
\end{proposition}

\begin{proposition}
\label{prop:2}
We have 
\begin{equation}
    \underset{x\in\Omega^c}{\inf}x^T\Lambda^{-1}x = \frac{1}{\max_{k}c_{k}^2\sigma_{k}^2}
\end{equation}
where we recall $\sigma_{k}^2=\rho_{k,k}$.
\end{proposition}

\begin{proof}[Proof of Proposition \ref{prop:1}]
\input{Proofs/Proposition-prop1}
\end{proof}

\begin{proof}[Proof of Proposition \ref{prop:2}]
\input{Proofs/Proposition-prop2}
\end{proof}

\begin{lemma}
\label{lem:2-2}
Let $M$ be a $J\times J$ square matrix and $j \in \{1,\dots, J\}$. Assume that $M_{-j,-j}$ and ${M_{j,j} - M_{j,-j}\cc{M_{-j,-j}}^{-1}M_{-j,j}}$ are invertible.
Then we have 
$$\pp{M^{-1}}_{j,j} = \pp{M_{j,j} - M_{j,-j}\cc{M_{-j,-j}}^{-1}M_{-j,j}}^{-1}.$$

\end{lemma}

\begin{proof}[Proof of Lemma \ref{lem:2-2}]
\input{Proofs/Lemma-2-2}
\end{proof}

%% file: Proofs/Proposition-prop1.tex
Letting $Z_k^n =  \mathbb{P}\pp{\hat{\theta}_n = \thetaq^n_k \mid \rz_1}$ and $p_k^n =  \mathbb{P}\pp{\hat{\theta}_n = \thetaq^n_k}$, we have
\begin{align*}
  \|\mathbb{P}_{(\hat{\theta}_n,\rz_0)}-\mathbb{P}_{(\hat{\theta}_n,\rz_1)}\|_{TV} &= \frac{1}{2}\sum_{k=1}^{K}\mathbb{E}\left[\left| p_k^n - Z_k^n\right|\right] \\ 
    &\leq \frac{1}{2}\mathbb{E}\left[\left|p_1^n - Z_1^n\right|\right]  + \frac{1}{2}\sum_{k=2}^{K}\mathbb{E}\left[ p_k^n +Z_k^n\right] \\ 
    &= \frac{1}{2}\mathbb{E}\left[\left|(1-p_1^n) - (1-Z_1^n)\right|\right]  + \sum_{k=2}^{K} p_k^n \\ 
    &\leq \frac{1}{2}\mathbb{E}\left[(1-p_1^n) + (1-Z_1^n)\right]  + (1- p_1^n)\\ 
    &= 2(1- p_1^n).
\end{align*}
Recall that $\gA_{\gQ_n}$ minimizes the empirical loss. Letting $\Omega_{K-1} = [-1,\infty)^{K-1}$, we then have 
\begin{align}
\notag
    p_1^n &= \mathbb{P}\pp{1 = \underset{k}{\argmin}\left\{\frac{1}{n}\sum_{j=1}^{n} \ell\pp{\thetaq_k,\rz_j}\right\}} \\ 
\notag
    &=\mathbb{P}\pp{\forall k > 1, \frac{1}{n}\sum_{j=1}^{n}\cc{\ell\pp{\thetaq_k,\rz_j} - \ell\pp{\thetaq_1,\rz_j}}\geq 0} \\ 
\notag
    &= \mathbb{P}\pp{\forall k > 1, \frac{1}{n}\sum_{j=1}^n \cc{L_{k,j} - \delta_k} \geq - \delta_k} \\ 
\label{eq:Prob}
    &= \mathbb{P}\pp{\frac{1}{n}\sum_{j=1}^{n}\cc{L_{k,j} - \delta_k}_{k>1} \in D\Omega_{K-1}} \\ 
\notag
    &=\mathbb{P}\pp{\frac{1}{n}\sum_{j=1}^{n}D^{-1}\cc{L_{k,j} - \delta_k}_{k>1} \in \Omega_{K-1}},
\end{align}

which gives

\begin{align*}
    1 - p_1^n &= \mathbb{P}\pp{\frac{1}{n}\sum_{j=1}^{n}D^{-1}\cc{L_{k,j} - \delta_k}_{k>1} \in \Omega_{K-1}^c},
\end{align*}
so that
\begin{align*}
   \|\mathbb{P}_{(\hat{\theta}_n,\rz_0)}-\mathbb{P}_{(\hat{\theta}_n,\rz_1)}\|_{TV} &\leq 2\mathbb{P}\pp{\frac{1}{n}\sum_{j=1}^{n}\cc{D^n}^{-1}\cc{L_{k,j}^n - \mkn{k}}_{k>1} \in \Omega_{K-1}^c}\\
    &= 2\underbrace{\mathbb{P}\pp{\frac{1}{n\mkn{2}}\sum_{j=1}^{n}\cc{\pp{\mkn{2}}^{-1}D^n}^{-1}\cc{L_{k,j}^n - \mkn{k}}_{k>1} \in \Omega_{K-1}^c}}_{1-p_1^n}.
\end{align*}

By {\bf(A2)}, $\cc{\pp{\mkn{2}}^{-1}D^n}^{-1}\cc{L_{k,j}^n - \mkn{k}}_{k>1}$ lives (asymptotically) in a $(K-1)$-dimensional euclidean sub-space. By {\bf(A3)} and {\bf (A4)}, as we live in an Hilbert space,  using \cite{araujo1980central}, we have $\gL\pp{\frac{1}{\sqrt{n}}\sum_{j=1}^n \cc{\pp{\mkn{2}}^{-1}D^n}^{-1}\cc{L_{k,j}^n - \mkn{k}}_{k>1} } \toinf{n} \gamma\coloneqq \gN_{K-1}\pp{0, \Lambda}$, where $\gN_{K-1}$ is the $(K-1)$-dimensional Gaussian distribution.
Now, using the fact that $\E\cc{L_{k,j}^n} = \delta_k^n$, using {\bf (A3)} and {\bf (A4)} and the convergence to a Gaussian measure, we apply Theorem 2.2 of \cite{de1992moderate} to get

\begin{align*}
   \limsup_{n\to \infty} \frac{1}{n\pp{\mkn{2}}^2}\log \pp{1-p_1^n} &\leq -\underset{x\in\Omega_{K-1}^c}{\inf}\left\{
   \begin{array}{ll}
       \frac{\displaystyle x^T\Lambda^{-1}x}{2}  & \text{, if }x\in H_{\gamma}\\
       \infty & \text{, otherwise}
   \end{array}
   \right.\\ 
   &= - \underset{x\in\Omega^c}{\inf}\; \frac{x^T\Lambda^{-1}x}{2},
\end{align*}

where $H_\gamma$ is the Hilbert space associated with $\gamma$ (see \cite{de1992moderate}). Additionally, as $\sqrt{n}\delta_2^n\toinf{n}\infty$ by {\bf (A1)}, we have $\frac{1}{n(\delta_2^n)^2} \log 2 \toinf{n} 0$.

%% file: Proofs/Proposition-prop2.tex
Let $M=\Lambda^{-1}$. Note that $\Omega^c = \{x\in\R^{K-1} : \exists j, x_j < -1\}$, giving

$$\underset{x\in\Omega^c}{\inf}\; x^TM x = \underset{j}{\min}\underset{x_{-j}\in\R^{K-1}}{\inf}\underset{x_j<-1}{\inf} x^TMx,$$

where we write $x_{-j}$ equals $x$ where we omit its $j^{th}$ entry. We then shall write 

\begin{align*}
    x^TMx &= x_j^2M_{j,j} + 2x_jx_{-j}^T M_{-j,j} + x_{-j}^TM_{-j,-j}x_{-j}
\end{align*}
where $M_{-j,-j}$ is the sub-matrix of $M$ consisting of all entries except the $j^{th}$ row and column and $M_{-j,j}$ is the $j^{th}$ column of $M$ except its $j^{th}$ entry. 

The infimum must be reached on the frontier of the set, i.e. such that $x_j=-1$ for some $j$. Indeed, assuming $x$ in the interior of $\Omega^c$, we have that $x_j<-1$ for some $j$. Then for any $1<\alpha<|x_j|$, $x_j/\alpha < -1$ meaning that $x/\alpha$ still belongs to the interior of $\Omega^c$. However, $(x/\alpha)^TM(x/\alpha) = \frac{1}{\alpha^2}x^TMx < x^TMx$, which shows that $x$ was not optimal. For an optimal $x$, we then have 

\begin{align*}
    x^TMx &= M_{j,j} - 2x_{-j}^T M_{-j,j} + x_{-j}^TM_{-j,-j}x_{-j}.
\end{align*}

It is then sufficient to study the optimization problem over $x_{-j}$, which amounts down to the optimization of a quadratic function, whose minimum is then reached for $x_{- j}$ satisfying 

\begin{align*}
    \nabla_{x_{-j}}\pp{- 2x_{-j}^T M_{-j,j} + x_{-j}^TM_{-j,-j}x_{-j}} & = 0 \\ 
    \iff x_{-j} &= \pp{M_{-j,-j}}^{-1}M_{-j,j},
\end{align*}

giving

\begin{align*}
    \underset{x_{-j}\in\R^{J-1}}{\inf}\underset{x_j<-1}{\inf} x^TMx &= M_{j,j} - M_{j,-j}\pp{M_{-j,-j}}^{-1}M_{-j,j}.
\end{align*}

Applying Lemma \ref{lem:2-2} concludes the proof.

%% file: Proofs/Lemma-2-2.tex
First note that if $M$ is designed by block as follows,

\begin{align*}
    M&=\begin{pmatrix}
    A & B \\
    C & D 
    \end{pmatrix},
\end{align*}

then we have

\begin{equation}
\label{eq:M_inv}
    M^{-1}=\begin{pmatrix}
    \pp{A - BD^{-1}C}^{-1} & * \\
    * & * 
    \end{pmatrix},
\end{equation}

as long as $C$ and $A - BD^{-1}C$ are invertible.
Let $P_j$ and $Q_j$ be $J\times J$ matrices defined for all $j$ such that $P_j M$ permutes the first and the $j^{th}$ rows of $M$, and $Q_j M$ permutes the $(j-1)^{th}$ and the $j^{th}$ rows of $M$. For instance, if $J=4$ and $j=3$ then we have

\begin{align*}
    P_3&=\begin{pmatrix}
    0 & 0 & 1 & 0 \\ 
    0 & 1 & 0 & 0 \\ 
    1 & 0 & 0 & 0 \\ 
    0 & 0 & 0 & 1
    \end{pmatrix}
    ,&     Q_3&=\begin{pmatrix}
    1 & 0 & 0 & 0 \\
    0 & 0 & 1 & 0 \\
    0 & 1 & 0 & 0 \\
    0 & 0 & 0 & 1
    \end{pmatrix}.
\end{align*}

Note that we have $(P_j)^{-1} = P_j$ and $(Q_j)^{-1}=Q_j$. Let $R = Q_3Q_4\cdots Q_jP_jMP_jQ_j\cdots Q_4Q_3$. Developing the formula, we get 

\begin{equation}
\label{eq:RM}
R = \begin{pmatrix}
    M_{j,j} & M_{j,-j} \\ 
    M_{-j,j} & M_{-j,-j}
    \end{pmatrix}.
\end{equation}

On one hand, using \autoref{eq:M_inv} and \autoref{eq:RM}, we have $R^{-1} = \begin{pmatrix}
    \pp{M_{j,j} - M_{j,-j}\cc{M_{-j,-j}}^{-1}M_{-j,j}}^{-1}&\star \\ 
    \star&\star
    \end{pmatrix}$. On the other hand, distributing the inverse operator on the product of matrices and using again \autoref{eq:M_inv}, we have 

\begin{align*}
    R^{-1} &= Q_3Q_4\cdots Q_jP_jM^{-1}P_jQ_j\cdots Q_4Q_3 \\ &= \begin{pmatrix}
    \pp{M^{-1}}_{j,j}&\star \\ 
    \star&\star
    \end{pmatrix},
\end{align*}

which concludes the proof.

%% file: Appendix/XII-Quantizations.tex
\section{Complete Algorithm}

We provide below (\autoref{algo:full_procedure}) the complete algorithm used to estimate $\qcertif$, incorporating the learning procedure $\gA$ and the quantization function $\gQ$.

\label{app:algo}
\begin{algorithm}
    \caption{Estimation of $\qcertif$}
    \label{algo:full_procedure}
    \begin{algorithmic}[1]
        \STATE \textbf{Input:} A training dataset $\gD_n$, a validation dataset $\gD_{\textrm{val}}$, a learning procedure $\gA$, a quantizer $\gQ$, an initialized model $\theta$, a number of epochs $K$.
        \STATE \textbf{Output:} An estimate of $\qcertif$.
        \STATE \textbf{Initialization:} Set the list of all quantized loss $\Ls_{\textrm{val}} = \{\}$, of all average quantized losses $m_{\textrm{val}} = \{\}$, and of all variances $\sigma_{\textrm{val}}^2 = \{\}$.
        \FOR{$k=1$ to $K$}
            \STATE $\theta \leftarrow \gA(\theta, \gD_n)$.
            \STATE $\thetaq \leftarrow \gQ(\theta)$.
            \STATE $\Ls_k \leftarrow \{\ell(\thetaq, \rz) : \rz\in\gD_{\textrm{val}}\}$.
            \STATE $m_k \leftarrow \frac{1}{|\gD_{\textrm{val}}|}\sum_{\rz\in\gD_{\textrm{val}}}\ell(\thetaq, \rz)$.
            \STATE $\Ls_{\text{val}}[k] \leftarrow \Ls_k$.
        \ENDFOR
        \STATE $\text{idx} \leftarrow \text{argsort}(m_{\text{val}})$.
        \STATE $m_{\textrm{val}} \leftarrow m_{\textrm{val}}[\text{idx}]$.
        \STATE $\Ls_{\textrm{val}} \leftarrow \Ls_{\textrm{val}}[\text{idx}]$.
        \FOR{$k=2$ to K}
            \STATE $\sigma_{\textrm{val}}^2[k] \leftarrow \textrm{Var}\pp{\Ls_{\text{val}}[k] - \Ls_{\text{val}}[1]}$.
        \ENDFOR
        \STATE $r_{\gQ} \leftarrow \frac{1}{2}\cc{\underset{2\leq k\leq K}\max{\left(\sigma_{\text{val}}^2[k] \times \pp{\frac{m_{\text{val}}[2] - m_{\text{val}}[1]}{m_{\text{val}}[k] - m_{\text{val}}[1]}}^2\right)}}^{-1}$.
        \STATE \textbf{return} $\qcertif = r_{\gQ} \pp{m_{\text{val}}[2] - m_{\text{val}}[1]}^2$.

    \end{algorithmic}
\end{algorithm}

\section{Quantizations for the Synthetic and Molecular Experiments}
\label{app:quantization}

\begin{figure}
    \centering
    \includegraphics[width=0.9\linewidth]{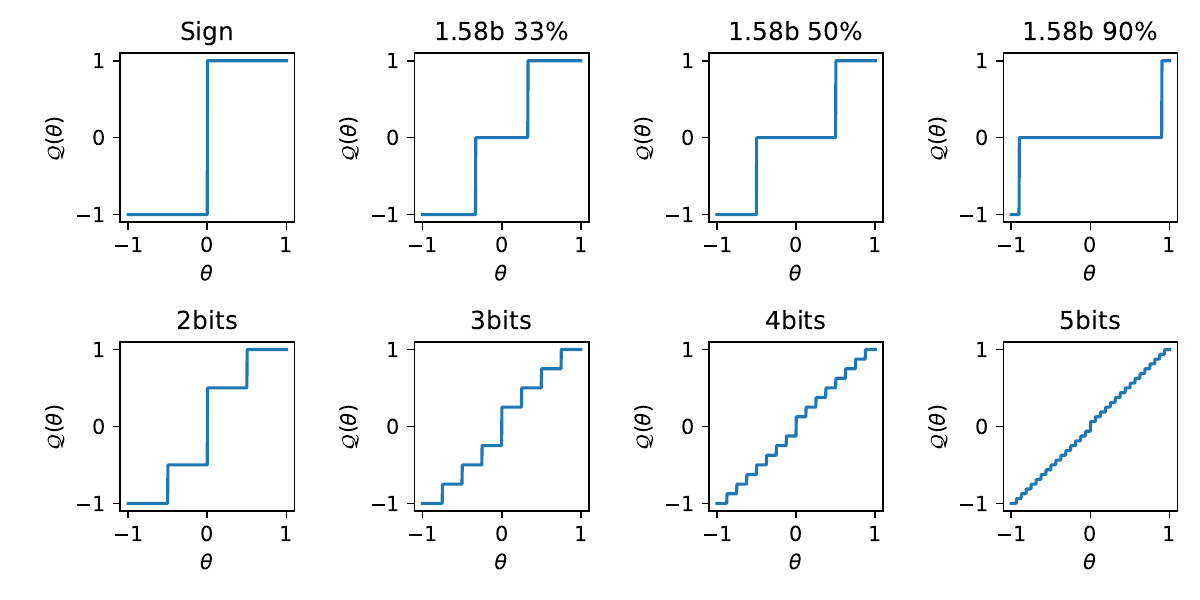}
    \caption{Illustration of the quantization functions used on the interval $[-1, 1]$.}
    \label{fig:quantizers}
\end{figure}

\begin{table*}
    \centering
    \caption{
        Description of the quantizers used in the experiments.
    }
    \label{tab:quantizers}
    \begin{center}
        \input{Tables/quantizers}
    \end{center}
\end{table*}

For our synthetic and molecular experiments, we used simple quantizers to facilitate the analysis of the results.
These quantizers are summarized in \autoref{tab:quantizers}, and~\autoref{fig:quantizers} illustrates how the different functions used quantize the interval $[-1, 1]$.

%% file: Tables/quantizers.tex
\begin{tabular}{c|c}
{quantizer} & {$\gQ(\theta_j)$} \\
\midrule
Sign &  $\gQ(\theta_j) = \frac{\theta_j}{|\theta_j|}$\\
\midrule
1.58b 33\% & $\gQ(\theta_j) = \left\{
   \begin{array}{ll}
        \frac{\theta_j}{|\theta_j|},& \text{if }|\theta_j| < q(|\theta|, 0.33)\\
       0, & \text{otherwise}
   \end{array}
   \right.$\\
\midrule
1.58b 50\%  & $\gQ(\theta_j) = \left\{
   \begin{array}{ll}
        \frac{\theta_j}{|\theta_j|},& \text{if }|\theta_j| < q(|\theta|, 0.5)\\
       0, & \text{otherwise}
   \end{array}
   \right.$\\
\midrule
1.58b 90\%  & $\gQ(\theta_j) = \left\{
   \begin{array}{ll}
        \frac{\theta_j}{|\theta_j|},& \text{if }|\theta_j| < q(|\theta|, 0.9)\\
       0, & \text{otherwise}
   \end{array}
   \right.$\\
\midrule
2 bits & $\gQ(\theta_j) =\frac{\theta_j}{|\theta_j|} \frac{\alpha}{2} \times \text{int}\pp{1+ \text{clip}(\frac{2\theta_j}{\alpha}, 0, 2)}, \quad \alpha = 2^{\text{round}(\log_2\pp{\max |\theta|})}$\\
\midrule
3 bit & $\gQ(\theta_j) =\frac{\theta_j}{|\theta_j|} \frac{\alpha}{4} \times \text{int}\pp{1+ \text{clip}(\frac{4\theta_j}{\alpha}, 0, 4)}, \quad \alpha = 2^{\text{round}(\log_2\pp{\max |\theta|})}$\\
\midrule
4 bits & $\gQ(\theta_j) =\frac{\theta_j}{|\theta_j|} \frac{\alpha}{8} \times \text{int}\pp{1+ \text{clip}(\frac{8\theta_j}{\alpha}, 0, 8)}, \quad \alpha = 2^{\text{round}(\log_2\pp{\max |\theta|})}$\\
\midrule
5 bits & $\gQ(\theta_j) =\frac{\theta_j}{|\theta_j|} \frac{\alpha}{16} \times \text{int}\pp{1+ \text{clip}(\frac{16\theta_j}{\alpha}, 0, 16)}, \quad \alpha = 2^{\text{round}(\log_2\pp{\max |\theta|})}$\\
\bottomrule
\end{tabular}

%% file: Appendix/XIII-Synthetic_Details.tex
\section{Baseline estimation of the MIS}
\label{app:baseline_estimation}
We describe here the baseline approach used to estimate the membership inference score (MIS) of a model $\thetan$.

Inferring membership of $\Tilde{\rz}$ by an MIA can naturally be considered as a statistical test,
\begin{equation*}
\left\{
    \begin{array}{ll}
  H_0: &``\Tilde{\rz}{\text{ belongs to the training dataset}}".  \\
  H_1: & ``\Tilde{\rz}{\text{ does not belong to the training dataset}}".
\end{array}\right.
\end{equation*}
Under the mathematical setting presented before, these hypotheses can be apprehended as deciding whether $\hat{\theta}_n$ is independent or not to $\Tilde{\rz}$. Specifically, testing $H_0$ against $H_1$ is equivalent to testing $H_0'$ against $H_1'$, where
\begin{equation}
\left\{
    \begin{array}{ll}
    H_0': & (\hat{\theta}_n, \Tilde{\rz})\sim P_{\pp{\hat{\theta}_n,\rz_1}}. \\
    H_1': & (\hat{\theta}_n,\Tilde{\rz})\sim P_{\hat{\theta}_n}\otimes P.
\end{array}\right. 
\end{equation}
For any dominating measure $\zeta$ on $P_{\pp{\hat{\theta}_n,\rz_1}}$ (and $P_{\hat{\theta}_n}\otimes P$), denoting by $f$ (resp. $g$) the density of  $P_{\pp{\hat{\theta}_n,\rz_1}}$ (resp. $P_{\hat{\theta}_n}\otimes P$) with respect to $\zeta$, we have that $\phi^*(\theta, z) = \mathbbm{1}\left\{\frac{f(\theta,z)}{g(\theta,z)} \geq1\right\}$ satisfies:
\begin{equation}
    \underset{\phi}{\sup}\;{\text{Acc}}_n(\phi; P,\gA ) = {\text{Acc}}_n(\phi^*; P,\gA ).
\end{equation}
\noindent The function $\phi^*$ is the Neyman-Pearson test for $H_0'$ against $H_1'$. This suggests that evaluating empirically the MIS of an algorithm amounts to training a discriminator and evaluating it.
The baseline approach therefore consists in training a binary classifier $\bsl: \Theta \times \mathbb{R}^d \times \mathbb{R} \rightarrow [0,1]$ to distinguish between samples $z$ sampled from the training set of a given $\thetan$ and sampled from the product distribution $P_{\thetan}\otimes P$, and evaluate its accuracy.
For a sample $z=(x,y)\sim P$, where $x\in \mathbb{R}^d$ is the input and $y$ its corresponding label, and a model $\thetan$, the discriminator minimizes the binary cross-entropy loss:
\[
    \ell_{\textrm{DISC}}(\thetan, z) = \text{BinaryCE}\left(\bsl(\thetan, x, y), \mathbbm{1}_{z\in \mathcal{D}_{\textrm{train}}(\thetan)}\right),
\]
where $\mathcal{D}_{\textrm{train}}(\thetan)$ is the training set of $\thetan$.

The discriminator is implemented as a feed-forward neural network that takes as input: $x$, the input data; the flattened parameters of $\thetan$; and the loss of the model $\thetan$ on $ x$ and $ y$.
For instance, if the model $\thetan$ is a binary classifier, $\bsl$ is a neural network with input $x\in \mathbb{R}^d$, $\thetan$ and the binary cross-entropy loss of $\thetan$ on $(x,y)$.

We train the discriminator using a set of models, $\{\thetan{_ i}\}_{i\in\{1,\ldots,k_{\text{run}}\}}$ where each $\thetan{_ i}$ is trained on an independent dataset $\{z_{i,1},\ldots, z_{i,n}\} \sim P$ of $n$ i.i.d. samples.
To generate negative samples, we independently sample additional sets, $z^{\textrm{neg}}_{i,1},\ldots,z^{\textrm{neg}}_{i,n} \sim P$, ensuring no overlap with the training sets of any models, or between the negative samples of different models.
This independence between datasets is crucial for preventing information leakage, but it may be restrictive in practice due to the high data requirements.

A key drawback of this approach arises when $\thetan$ contains multiple layers.
Different permutations of the weights of the hidden units can represent identical models, but the discriminator $\bsl$, which relies on flattened parameters, is not invariant to such permutations.

\FloatBarrier

\section{Synthetic experiments}
\label{app:synthetic_details}

\subsection{Training curves}

\begin{figure*}
    \centering
    \includegraphics[width=0.8\linewidth]{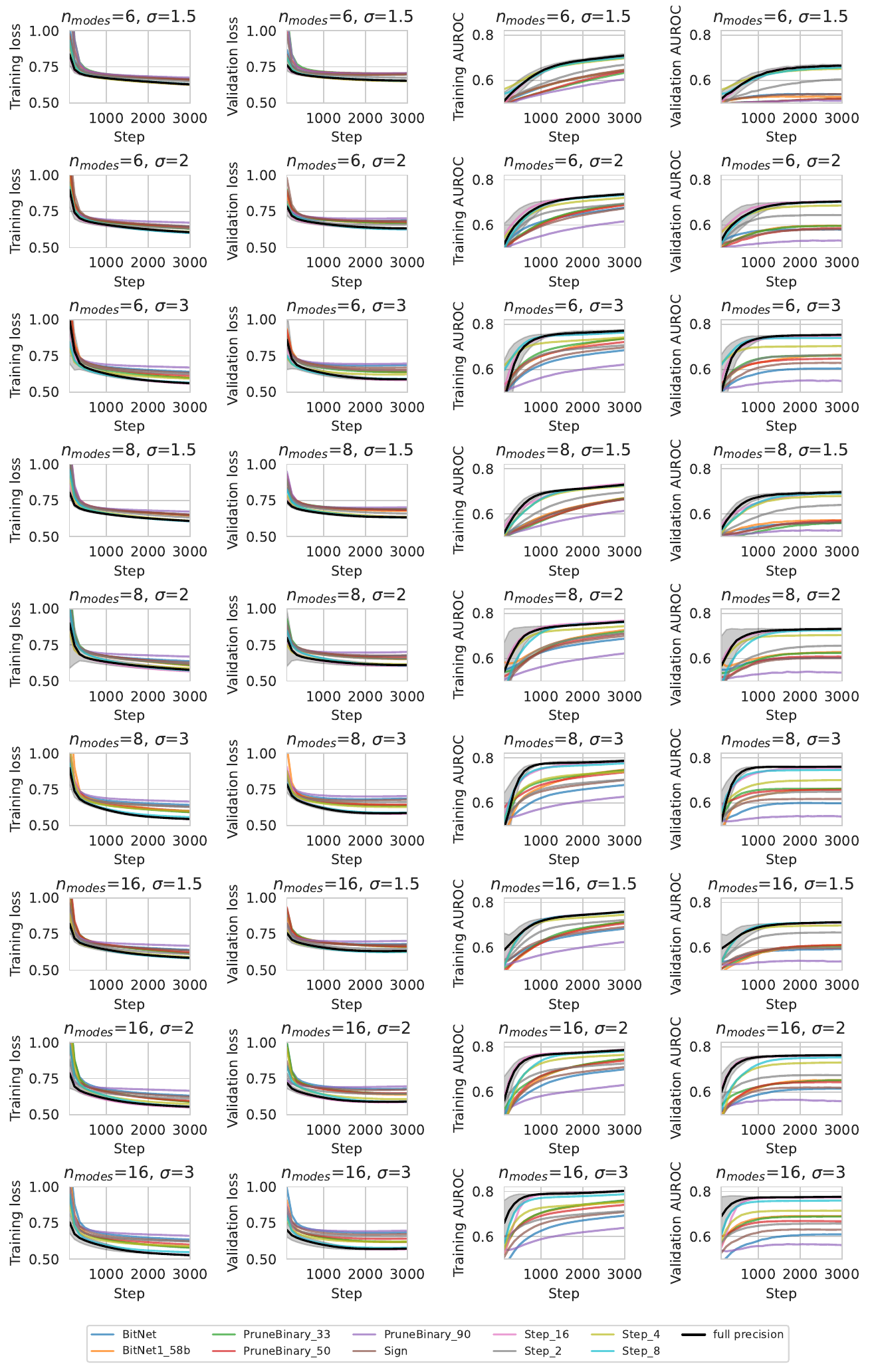}
    \caption{
        Training curves of the models trained on synthetic datasets, with the corresponding quantized metrics
    }
    \label{fig:synth_tc}
\end{figure*}

We provide in~\autoref{fig:synth_tc} the training curves of the classifiers trained on the synthetic datasets.

\subsection{Runtime comparison with the baseline approach}

As explained in~\autoref{app:baseline_estimation}, the baseline approach consists of training a discriminator to distinguish between samples from the training set of a given $\thetan$ and samples from the product distribution $P_{\thetan}\otimes P$.
Similarly, the $\qcertif$-based approach relies on the training of multiple models to average the values of $\qcertif$ obtained.

The computational overhead induced by the $\qcertif$-based approach, specifically computing the validation loss of the quantized models, is negligible compared to the total training time (1 second against 4 minutes).
Similarly, the training of the discriminator takes only about 40 minutes (all experiments were conducted on NVIDIA A6000 GPUs).

As a result, training multiple models $\thetan$ over multiple runs is the computational bottleneck of our privacy evaluations.
To properly evaluate the time required to obtain both rankings, one would have to answer the following question: `How many runs do I need to launch to ensure the ranking I obtained is stable?'

As discussed in the previous section, after $15$ runs, the rankings obtained with $\qcertif$ are already highly correlated with those obtained with 300 runs. In contrast, the rankings obtained with the baseline MIS method require $150$ runs to reach the same level of correlation.
As a result, the time required to obtain stable rankings with $\qcertif$ is significantly lower (\(\approx 1\)h) than with the baseline MIS method (\(\approx 10\)h).

\subsection{Visualization of the datasets}

\begin{figure*}[ht]
    \centering
    \begin{subfigure}{0.3\linewidth}
        \centering
        \includegraphics[width=\linewidth]{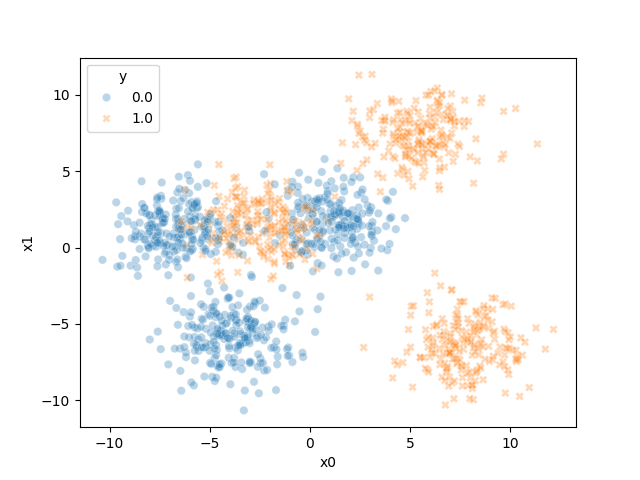}
        \caption{
            $n_{\textrm{cluster}} = 6, \sigma = 1.5$
        }
        \label{fig:6-1.5}
    \end{subfigure}
    \begin{subfigure}{0.3\linewidth}
        \centering
        \includegraphics[width=\linewidth]{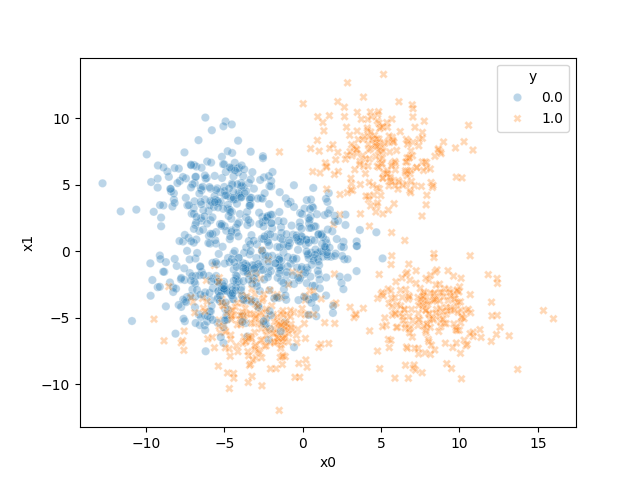}
        \caption{
            $n_{\textrm{cluster}} = 6, \sigma = 2$
        }
        \label{fig:6-2}
    \end{subfigure}
    \begin{subfigure}{0.3\linewidth}
        \centering
        \includegraphics[width=\linewidth]{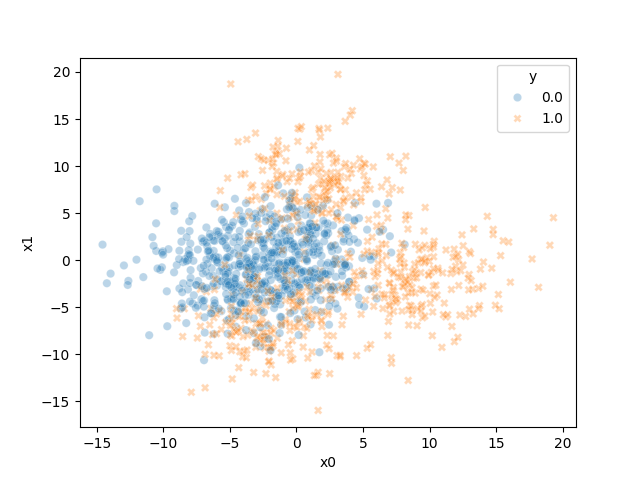}
        \caption{
            $n_{\textrm{cluster}} = 6, \sigma = 3$
        }
        \label{fig:6-3}
    \end{subfigure}
        \begin{subfigure}{0.3\linewidth}
        \centering
        \includegraphics[width=\linewidth]{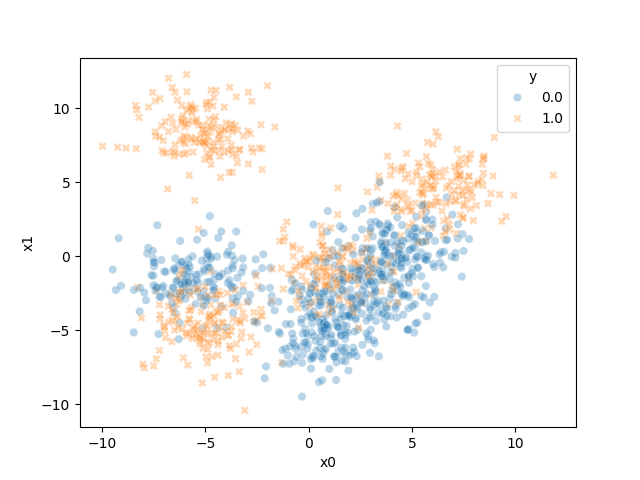}
        \caption{
            $n_{\textrm{cluster}} = 8, \sigma = 1.5$
        }
        \label{fig:8-1.5}
    \end{subfigure}
    \begin{subfigure}{0.3\linewidth}
        \centering
        \includegraphics[width=\linewidth]{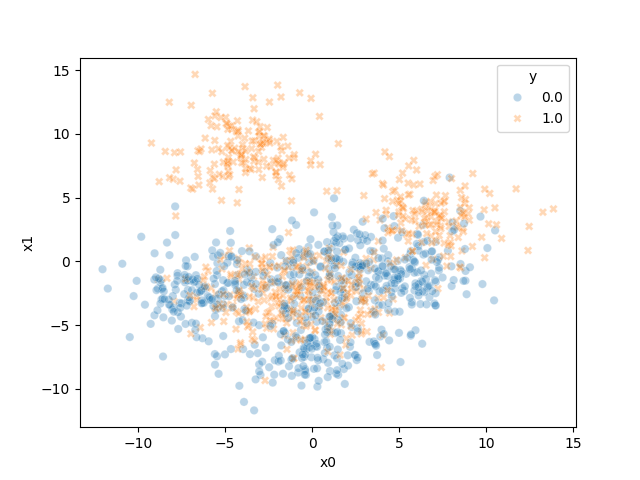}
        \caption{
            $n_{\textrm{cluster}} = 8, \sigma = 2$
        }
        \label{fig:8-2}
    \end{subfigure}
    \begin{subfigure}{0.3\linewidth}
        \centering
        \includegraphics[width=\linewidth]{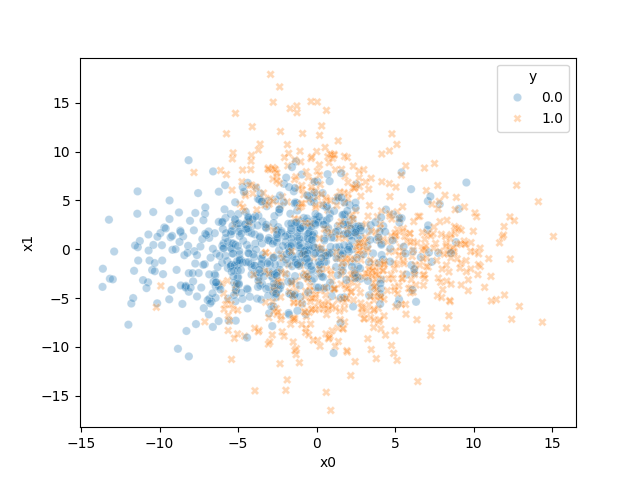}
        \caption{
            $n_{\textrm{cluster}} = 8, \sigma = 3$
        }
        \label{fig:8-3}
    \end{subfigure}
    \begin{subfigure}{0.3\linewidth}
        \centering
        \includegraphics[width=\linewidth]{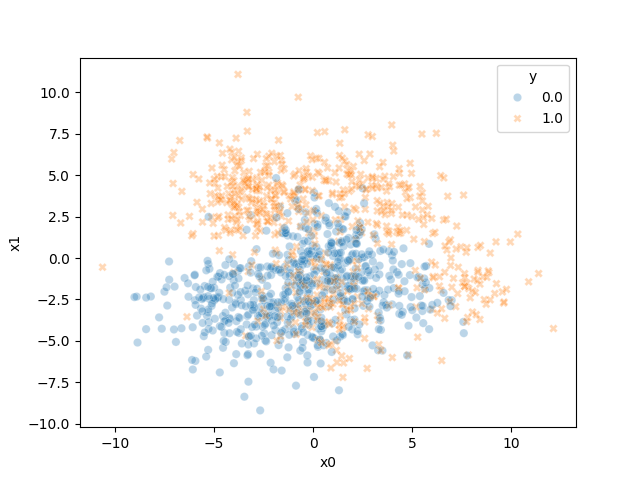}
        \caption{
            $n_{\textrm{cluster}} = 16, \sigma = 1.5$
        }
        \label{fig:16-1.5}
    \end{subfigure}
    \begin{subfigure}{0.3\linewidth}
        \centering
        \includegraphics[width=\linewidth]{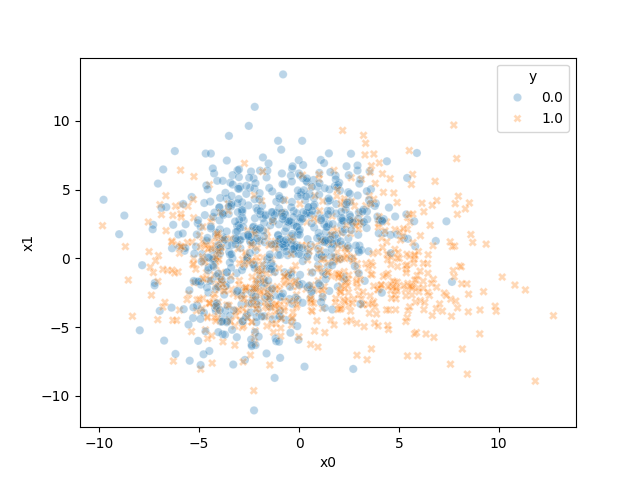}
        \caption{
            $n_{\textrm{cluster}} = 16, \sigma = 2$
        }
        \label{fig:16-2}
    \end{subfigure}
    \begin{subfigure}{0.3\linewidth}
        \centering
        \includegraphics[width=\linewidth]{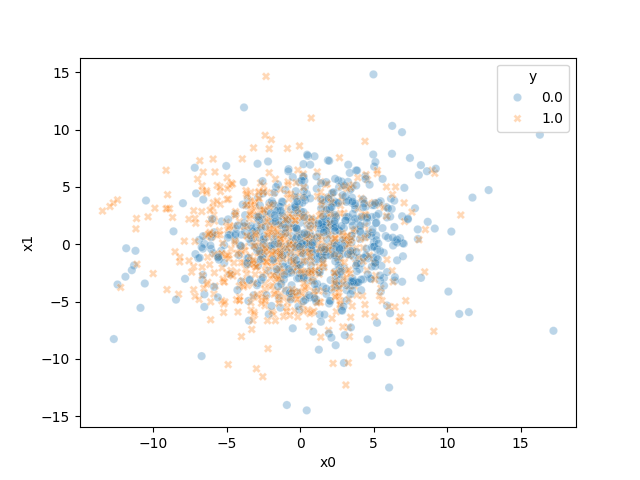}
        \caption{
            $n_{\textrm{cluster}} = 16, \sigma = 3$
        }
        \label{fig:16-3}
    \end{subfigure}
    \caption{
        Visualization of the synthetic datasets used in the experiments, through a PCA projection in dimension 2 (the original space is $\mathbb{R}^{128}$).
    }
    \label{fig:synth_datasets}
\end{figure*}

We provide in~\autoref{fig:synth_datasets} a visualization of the synthetic datasets used in the experiments, through a PCA projection in dimension 2.
This visualization helps understand how different data distribution might result in different empirical results, as some datasets are more challenging than others, such as the dataset with $n_{\textrm{cluster}} = 6$ and $\sigma = 1.5$, for whom the labels of the datapoints are easily separable, while $n_{\textrm{cluster}} = 16$ and $\sigma = 3$ provides a more challenging dataset, with overlapping clusters.

\FloatBarrier
\newpage

%% file: Appendix/XIV-NLP_Details.tex
\newpage

\section{Text Experiments}
\label{app:NLP}

\subsection{Training curves}

\begin{figure}[H]
    \centering
    \includegraphics[width=0.5\linewidth]{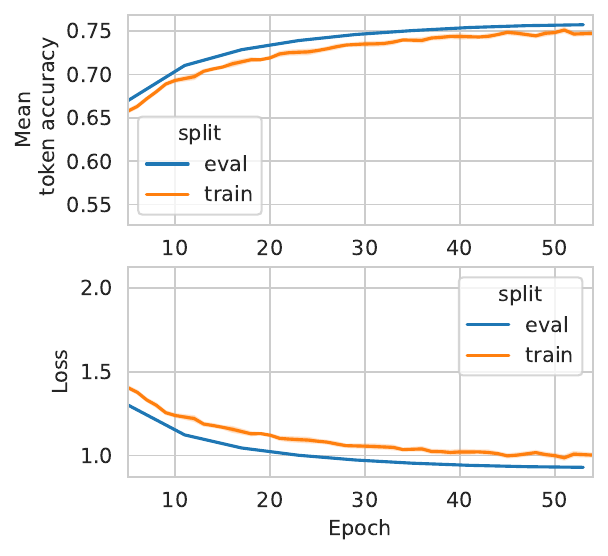}
    \caption{
        Training curves of the language models fine-tuned on SMolInstruct with a LORA adapter of rank 64.
    }
    \label{fig:nlp_tc}
\end{figure}

We provide the training curves on the language models fine-tuned for our NLP experiments in~\autoref{fig:nlp_tc}.

\subsection{Black-Box Attacks}

We used various black-box MIAs for LLMs on our language models. Each of them defines the probability of being a member of the model's training set as described below.
\begin{enumerate}
    \item LOSS~\cite{lossattack}: where the prediction of the MIA solely relies on the loss of the model  ($P(\textrm{member}) \propto -\mathcal{L}_{\theta}(x)$).
    \item min-k\%~\cite{shi2024detectingpretrainingdatalarge}: where the prediction of the MIA solely relies on the loss of 20\% tokens predicted with the lowest probability  ($P(\textrm{member}) \propto -\min_{J \subset \{1,..,N\}, |J|=\text{int}(|x|*0.2) }\sum_{j\in J} \mathcal{L}_{\theta}(x_i)$).
    \item LOSS + zlib~\cite{attacks}: where the loss is normalized with the size of the zlib compression of the text ($P(\textrm{member}) \propto -\mathcal{L}_{\theta}(x) / |zlib(x)|$).
    \item LOSS + ref~\cite{attacks}: where the loss of the model we attack is normalized by the loss of reference models (the ones obtained from the other runs)  ($P(\textrm{member}) \propto -(\mathcal{L}_{\theta}(x) - \sum_{i}\mathcal{L}_{\theta_{ref_i}}(x))$)
\end{enumerate}

Finally, these attacks can be combined (which we denote as LOSS + zlib + ref, for example).
In practice, we found that most attacks performed no better than a random attack, an observation consistent with previous works on black-box attacks with LLMs~\cite{duan2024do}. The only competitive attacks are those using the pool of reference models, which are the most expensive ones. We believe this is because our models did not overfit on their training set, making the direct observation of their loss insufficient to evaluate the membership of a sample.

\subsection{Additional Results}

\begin{figure*}
    \centering
    \includegraphics[width=.8\linewidth]{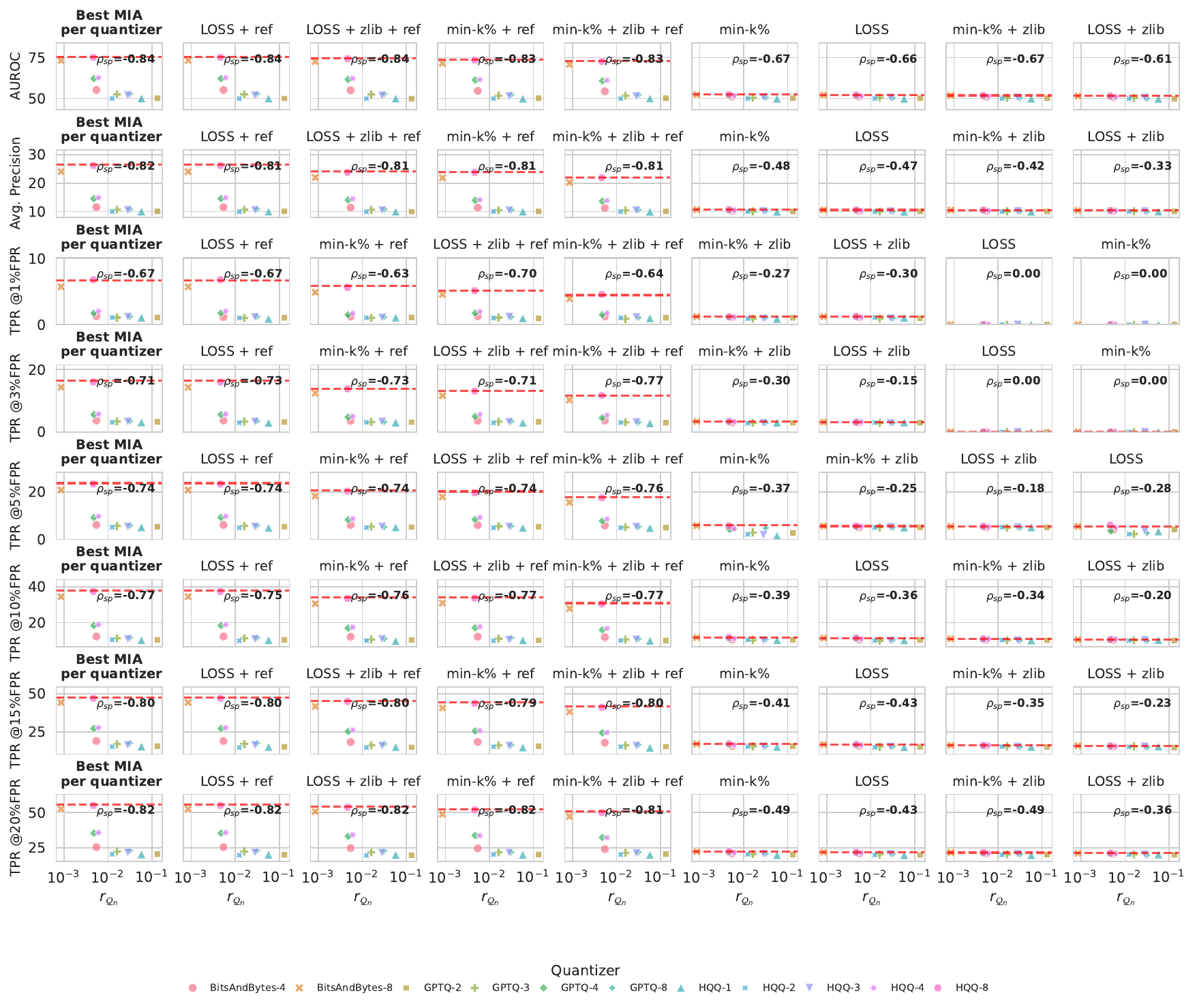}
    \caption{
        Quantized security against the performance of various MIAs for LLMs.
        The first figure displays the best metric achieved by an MIA for each quantizer against the quantized security $\qcertif$.
        We report the average-precision, auroc, and TPR at 20, 15, 10, 5, 3, and 1\% FPR.
    }
    \label{fig:all_metrics}
\end{figure*}

\begin{figure*}
    \centering
    \includegraphics[width=.8\linewidth]{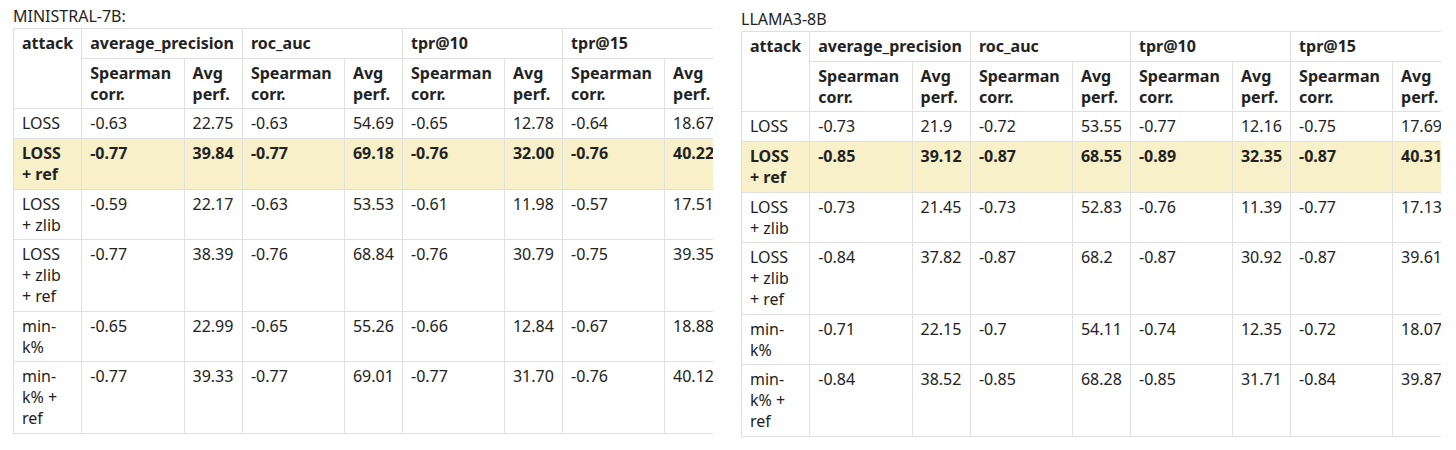}
    \caption{
        Average correlation between the quantized security $\qcertif$ and the efficiency of various MIAs for LLMs. The first table displays the result for MINISTRAL-7B and the second table for LLAMA3-8B. We report for the various MIAs, the correlation, averaged over different quantizers, between $\qcertif$ and the precision, auroc, TPR at 15 and 10 \% FPR.
    }
    \label{fig:big_llms}
\end{figure*}

We report in~\autoref{fig:all_metrics} the dependence of the quantized security $\qcertif$ with the best metric achieved by an MIA for each quantizer.
We provide results for: AUROC, average precision, TPR at 10, 5, and 1\% FPR.
Overall, our metric correlates with all the attacks, including the reference models, and we observe that the quantized security $\qcertif$ is a good indicator of the quantization robustness of the learning procedure against MIAs.
For other attacks, $\qcertif$ does not seem to correlate with their performances, which can be explained by the fact that these attacks perform close to random guesses.

%% file: Appendix/XV-Molecular_Details.tex
\FloatBarrier

\section{Molecular experiments}
\label{app:molecular_details}

\subsection{Experimental setup and Comprehensive results}
\label{subsec:comp_results}

\begin{figure}[H]
    \centering
    \includegraphics[width=0.75\linewidth]{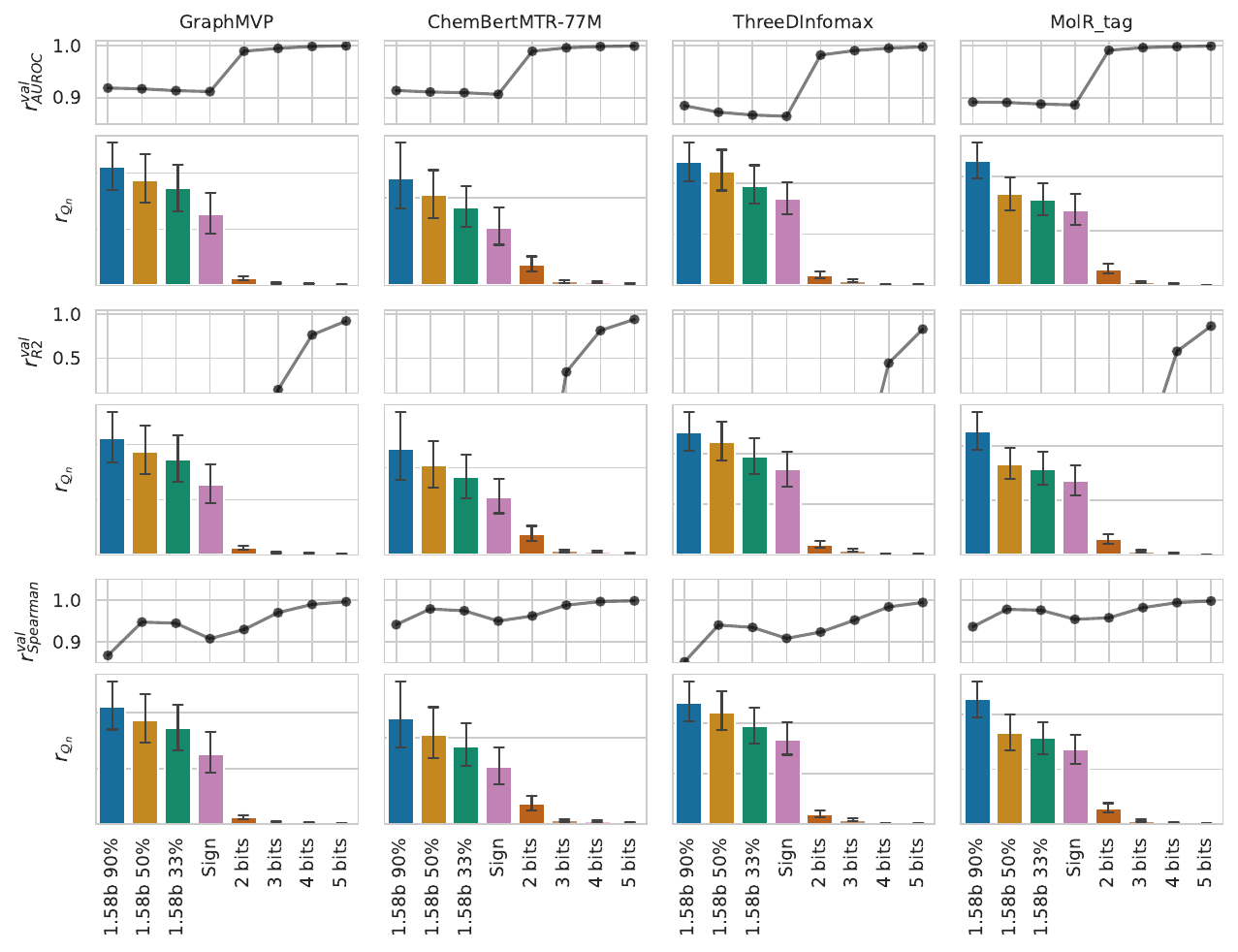}
    \caption{
        \textbf{Impact of quantization on classification tasks.}
        Evolution of the privacy of each downstream model $\qcertif$ along with relative performances of the quantized models compared to the original on classification/regression tasks.
    }
    \label{fig:barplot_dperfs_vs_rdelta_cls_dataset}
\end{figure}

We show in~\autoref{tab:mol_classification},~\autoref{tab:mol_reg}, and \autoref{fig:barplot_dperfs_vs_rdelta_cls_dataset} the comprehensive results of the quantized models on the classification and regression tasks, respectively.

We observe that while the quantized models are generally less accurate than the original models, they still achieve reasonable performance on the classification tasks.
In regression examples, the performance of quantized models is significantly lower than that of the original models.
In particular, when the quantization is on less than 4 bits, the prediction of the molecular properties is almost consistently lower than a simple mean prediction.
As explained in~\autoref{subsec:molecular_expe}, this result is expected, as while classification tasks rely on the definition of the boundary between classes, regression tasks require a fine-grained prediction of the target value.
However, while the direct predictions of the quantized models do not provide a good estimate of the target value, the ordering of the predictions is still preserved, as shown by the Spearman correlation between the quantized models' predictions and the labels in~\autoref{tab:mol_reg_sp} and~\autoref{fig:barplot_dperfs_vs_rdelta_cls_dataset}.

\begin{table*}
    \caption{
        AUROC performance of the quantized models on the classification tasks, averaged over all embedders.
    }\label{tab:mol_classification}
    \begin{center}
    \resizebox{\textwidth}{!}{
        \input{Tables/mol/mol_classification}
    }
    \end{center}
\end{table*}

\begin{table*}
    \caption{
        R2 performance of the quantized models on the regression tasks, averaged over all embedders. If the R2 score is less than -1, we display -inf for clarity.
    }
    \label{tab:mol_reg}
    \begin{center}
        \resizebox{\textwidth}{!}{
            \input{Tables/mol/mol_reg}
        }
    \end{center}
\end{table*}

\begin{table*}
    \caption{
        Spearman correlations between the labels and the predictions of the quantized models on the regression tasks, averaged over all embedders.
    }
    \label{tab:mol_reg_sp}
    \begin{center}
        \resizebox{\textwidth}{!}{
                \input{Tables/mol/mol_reg_sp}
        }
    \end{center}
\end{table*}

\subsection{Details on the evaluation of the quantizers' privacy}
\label{subsec:eval_privacy}

\begin{figure}
    \centering
    \begin{subfigure}{0.47\textwidth}
        \centering
        \includegraphics[width=\linewidth]{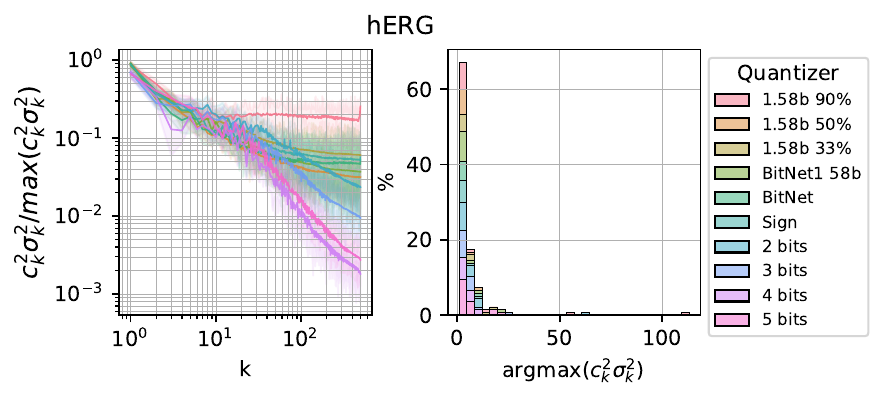}
    \end{subfigure}
    \begin{subfigure}{0.47\textwidth}
        \centering
        \includegraphics[width=\linewidth]{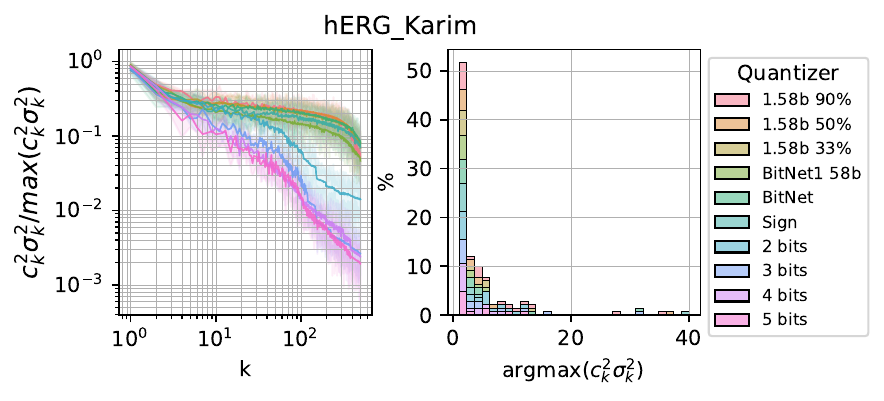}
    \end{subfigure}
    \begin{subfigure}{0.47\textwidth}
        \centering
        \includegraphics[width=\linewidth]{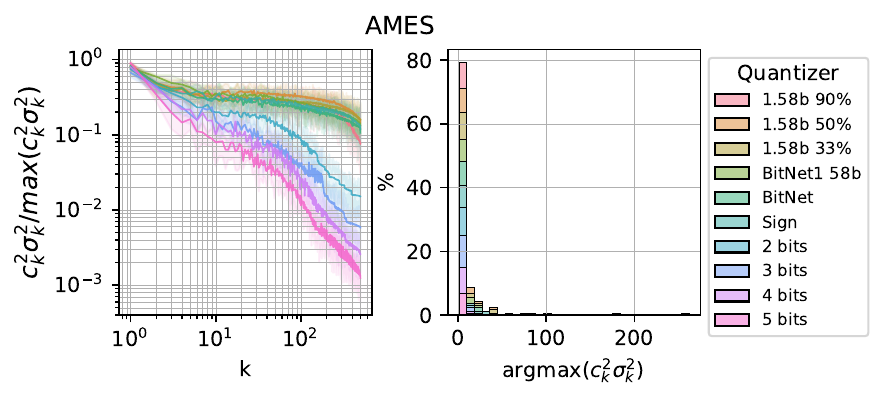}
    \end{subfigure}
    \begin{subfigure}{0.47\textwidth}
        \centering
        \includegraphics[width=\linewidth]{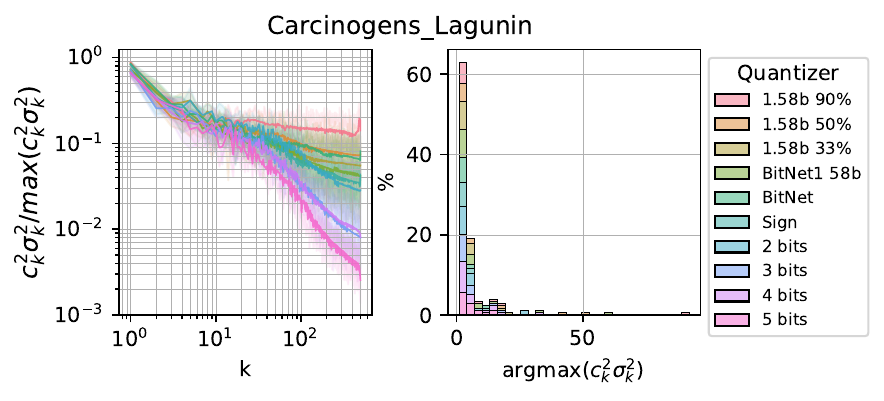}
    \end{subfigure}
    \begin{subfigure}{0.47\textwidth}
        \centering
        \includegraphics[width=\linewidth]{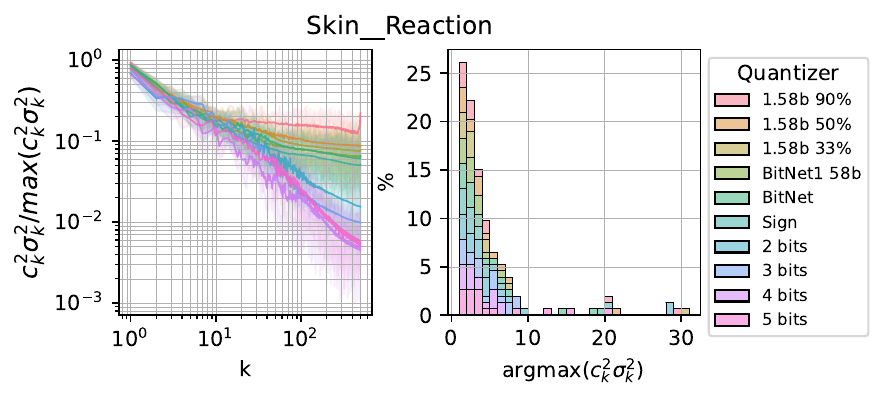}
    \end{subfigure}
    \begin{subfigure}{0.47\textwidth}
        \centering
        \includegraphics[width=\linewidth]{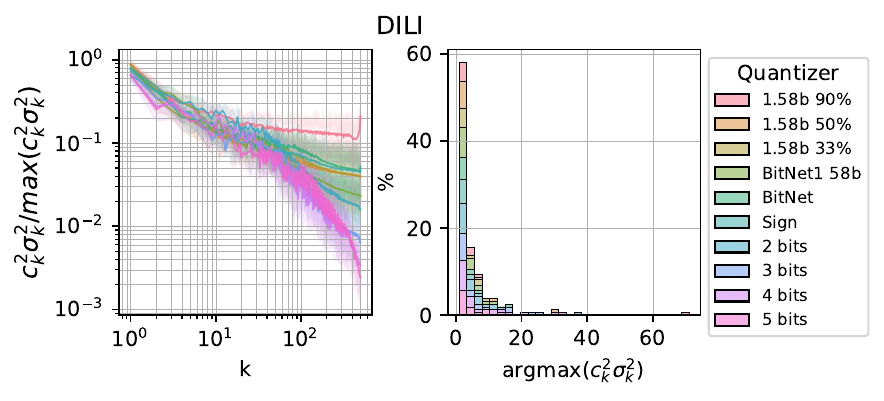}
    \end{subfigure}
    \caption{
        Evolution of $\Lambda_{k,k}$ with the k, where the indices are sorted with decreasing values of average loss, and the histogram of the index of the maximum value of $\Lambda_{k,k}$ for each quantizer.
    }
    \label{fig:k_max}
\end{figure}

Our hypothesis in the estimation of $\qcertif$  is that the quantized weights with the lowest average loss dominate the maximum value of $\lambda_k = \Lambda_{k,k} = \liminf{n}\pp{\mkn{2}/\mkn{k}}^2(\sigma_k^n)^2$.
We show in~\autoref{fig:k_max} the evolution of $\Lambda_{k,k}$ with the k, where the indexes are sorted with decreasing values of average loss, and the histogram of the index of the maximum value of $\Lambda_{k,k}$ for each quantizer, on four different datasets (trained on 500 epochs, hence $k \leq 500$).

The maximum value of $\Lambda_{k,k}$ is consistently reached for low $k$ values, which seems to confirm our hypothesis that the estimation of $\qcertif$ mainly revolves around the low loss values achieved by the quantizer, thereby validating our sampling strategy for estimating $\qcertif$.

%% file: Tables/mol/mol_classification.tex
\begin{tabular}{r|cccccccc|c}
{} & {\rotatebox{90}{\shortstack{Sign}}} & {\rotatebox{90}{\shortstack{1.58b 33\%}}} & {\rotatebox{90}{\shortstack{1.58b 50\%}}} & {\rotatebox{90}{\shortstack{1.58b 90\%}}} & {\rotatebox{90}{\shortstack{2 bits}}} & {\rotatebox{90}{\shortstack{3 bits}}} & {\rotatebox{90}{\shortstack{4 bits}}} & {\rotatebox{90}{\shortstack{5 bits}}} & {\rotatebox{90}{\shortstack{original}}} \\
{dataset} & {} & {} & {} & {} & {} & {} & {} & {} & {} \\
\midrule
AMES & 0.773 \tiny (89\%) & 0.781 \tiny (90\%) & 0.784 \tiny (90\%) & 0.748 \tiny (86\%) & 0.843 \tiny (97\%) & 0.853 \tiny (98\%) & 0.859 \tiny (99\%) & \textbf{0.861 \tiny (99\%)} & \textbf{\underline{0.862 \tiny (100\%)}} \\
BBB Martins & 0.800 \tiny (89\%) & 0.803 \tiny (89\%) & 0.807 \tiny (90\%) & 0.804 \tiny (89\%) & 0.890 \tiny (99\%) & \textbf{0.895 \tiny (99\%)} & \textbf{0.895 \tiny (99\%)} & \textbf{\underline{0.896 \tiny (100\%)}} & \textbf{\underline{0.896 \tiny (100\%)}} \\
Bioavailability Ma & 0.586 \tiny (94\%) & 0.588 \tiny (94\%) & 0.590 \tiny (95\%) & 0.579 \tiny (93\%) & 0.619 \tiny (99\%) & \textbf{0.622 \tiny (100\%)} & \textbf{0.622 \tiny (100\%)} & \textbf{\underline{0.623 \tiny (100\%)}} & \textbf{0.622 \tiny (100\%)} \\
CYP2C9 Substrate CarbonMangels & 0.558 \tiny (86\%) & 0.557 \tiny (86\%) & 0.561 \tiny (86\%) & 0.589 \tiny (90\%) & 0.642 \tiny (99\%) & 0.646 \tiny (99\%) & \textbf{0.647 \tiny (99\%)} & \textbf{0.647 \tiny (99\%)} & \textbf{\underline{0.648 \tiny (100\%)}} \\
CYP2C9 Veith & 0.787 \tiny (89\%) & 0.793 \tiny (90\%) & 0.799 \tiny (91\%) & 0.827 \tiny (94\%) & 0.868 \tiny (99\%) & 0.873 \tiny (99\%) & \textbf{0.876 \tiny (99\%)} & \textbf{\underline{0.877 \tiny (99\%)}} & \textbf{\underline{0.877 \tiny (100\%)}} \\
Carcinogens Lagunin & 0.766 \tiny (91\%) & 0.765 \tiny (91\%) & 0.772 \tiny (92\%) & 0.791 \tiny (94\%) & 0.824 \tiny (98\%) & 0.830 \tiny (99\%) & 0.831 \tiny (99\%) & \textbf{0.832 \tiny (99\%)} & \textbf{\underline{0.833 \tiny (100\%)}} \\
ClinTox & 0.561 \tiny (80\%) & 0.555 \tiny (79\%) & 0.557 \tiny (79\%) & 0.589 \tiny (84\%) & 0.698 \tiny (98\%) & 0.702 \tiny (99\%) & 0.704 \tiny (99\%) & \textbf{0.706 \tiny (99\%)} & \textbf{\underline{0.707 \tiny (100\%)}} \\
DILI & 0.827 \tiny (92\%) & 0.830 \tiny (93\%) & 0.831 \tiny (93\%) & 0.843 \tiny (94\%) & 0.888 \tiny (99\%) & \textbf{0.891 \tiny (99\%)} & \textbf{\underline{0.892 \tiny (99\%)}} & \textbf{\underline{0.892 \tiny (99\%)}} & \textbf{\underline{0.892 \tiny (100\%)}} \\
HIA Hou & 0.805 \tiny (91\%) & 0.805 \tiny (91\%) & 0.804 \tiny (91\%) & 0.790 \tiny (89\%) & \textbf{0.882 \tiny (99\%)} & \textbf{\underline{0.883 \tiny (99\%)}} & \textbf{\underline{0.883 \tiny (99\%)}} & \textbf{\underline{0.883 \tiny (100\%)}} & \textbf{\underline{0.883 \tiny (100\%)}} \\
PAMPA NCATS & 0.585 \tiny (81\%) & 0.583 \tiny (81\%) & 0.584 \tiny (81\%) & 0.583 \tiny (81\%) & 0.708 \tiny (99\%) & 0.711 \tiny (99\%) & \textbf{0.713 \tiny (99\%)} & \textbf{\underline{0.714 \tiny (99\%)}} & \textbf{\underline{0.714 \tiny (100\%)}} \\
Pgp Broccatelli & 0.856 \tiny (92\%) & 0.858 \tiny (92\%) & 0.859 \tiny (92\%) & 0.864 \tiny (93\%) & 0.921 \tiny (99\%) & \textbf{0.924 \tiny (99\%)} & \textbf{0.924 \tiny (100\%)} & \textbf{0.924 \tiny (100\%)} & \textbf{\underline{0.925 \tiny (100\%)}} \\
Skin  Reaction & 0.664 \tiny (89\%) & 0.667 \tiny (89\%) & 0.668 \tiny (90\%) & 0.670 \tiny (90\%) & 0.735 \tiny (99\%) & 0.740 \tiny (99\%) & 0.741 \tiny (99\%) & \textbf{0.742 \tiny (99\%)} & \textbf{\underline{0.743 \tiny (100\%)}} \\
hERG & 0.728 \tiny (91\%) & 0.726 \tiny (91\%) & 0.726 \tiny (91\%) & 0.739 \tiny (92\%) & 0.793 \tiny (99\%) & 0.795 \tiny (99\%) & \textbf{0.796 \tiny (99\%)} & \textbf{\underline{0.797 \tiny (99\%)}} & \textbf{\underline{0.797 \tiny (100\%)}} \\
hERG (k) & 0.767 \tiny (88\%) & 0.783 \tiny (90\%) & 0.789 \tiny (91\%) & 0.757 \tiny (87\%) & 0.819 \tiny (94\%) & 0.837 \tiny (96\%) & 0.855 \tiny (98\%) & \textbf{0.862 \tiny (99\%)} & \textbf{\underline{0.866 \tiny (100\%)}} \\
\end{tabular}

%% file: Tables/mol/mol_reg.tex
\begin{tabular}{r|cccccccc|c}
{} & {\rotatebox{90}{\shortstack{Sign}}} & {\rotatebox{90}{\shortstack{1.58b 33\%}}} & {\rotatebox{90}{\shortstack{1.58b 50\%}}} & {\rotatebox{90}{\shortstack{1.58b 90\%}}} & {\rotatebox{90}{\shortstack{2 bits}}} & {\rotatebox{90}{\shortstack{3 bits}}} & {\rotatebox{90}{\shortstack{4 bits}}} & {\rotatebox{90}{\shortstack{5 bits}}} & {\rotatebox{90}{\shortstack{original}}} \\
{dataset} & {} & {} & {} & {} & {} & {} & {} & {} & {} \\
\midrule
Caco2 Wang & -inf \tiny (-inf\%) & -inf \tiny (-inf\%) & -inf \tiny (-inf\%) & -inf \tiny (-inf\%) & -inf \tiny (-364\%) & 0.008 \tiny (1\%) & 0.458 \tiny (75\%) & \textbf{0.567 \tiny (93\%)} & \textbf{\underline{0.609 \tiny (100\%)}} \\
HydrationFreeEnergy FreeSolv & -inf \tiny (-inf\%) & -inf \tiny (-inf\%) & -inf \tiny (-inf\%) & -inf \tiny (-inf\%) & -0.466 \tiny (-70\%) & 0.412 \tiny (55\%) & 0.669 \tiny (91\%) & \textbf{0.715 \tiny (98\%)} & \textbf{\underline{0.725 \tiny (100\%)}} \\
LD50 Zhu & -inf \tiny (-inf\%) & -inf \tiny (-inf\%) & -inf \tiny (-inf\%) & -inf \tiny (-inf\%) & -inf \tiny (-578\%) & -0.129 \tiny (-25\%) & 0.339 \tiny (67\%) & \textbf{0.454 \tiny (89\%)} & \textbf{\underline{0.505 \tiny (100\%)}} \\
Lipophilicity (az) & -inf \tiny (-inf\%) & -inf \tiny (-inf\%) & -inf \tiny (-inf\%) & -inf \tiny (-inf\%) & -inf \tiny (-539\%) & -0.037 \tiny (-7\%) & 0.404 \tiny (72\%) & \textbf{0.508 \tiny (91\%)} & \textbf{\underline{0.552 \tiny (100\%)}} \\
PPBR AZ & -inf \tiny (-inf\%) & -inf \tiny (-inf\%) & -inf \tiny (-inf\%) & -inf \tiny (-inf\%) & -inf \tiny (-inf\%) & -0.480 \tiny (-233\%) & 0.022 \tiny (2\%) & \textbf{0.156 \tiny (65\%)} & \textbf{\underline{0.229 \tiny (100\%)}} \\
Solubility AqSolDB & -inf \tiny (-inf\%) & -inf \tiny (-inf\%) & -inf \tiny (-inf\%) & -inf \tiny (-inf\%) & -inf \tiny (-792\%) & -0.081 \tiny (-10\%) & 0.575 \tiny (72\%) & \textbf{0.740 \tiny (93\%)} & \textbf{\underline{0.792 \tiny (100\%)}} \\
VDss Lombardo & -inf \tiny (-inf\%) & -inf \tiny (-inf\%) & -inf \tiny (-inf\%) & -inf \tiny (-inf\%) & -0.859 \tiny (-287\%) & -0.010 \tiny (6\%) & 0.175 \tiny (73\%) & \textbf{0.224 \tiny (91\%)} & \textbf{\underline{0.248 \tiny (100\%)}} \\
\end{tabular}

%% file: Tables/mol/mol_reg_sp.tex
\begin{tabular}{r|cccccccc|c}
{} & {\rotatebox{90}{\shortstack{Sign}}} & {\rotatebox{90}{\shortstack{1.58b 33\%}}} & {\rotatebox{90}{\shortstack{1.58b 50\%}}} & {\rotatebox{90}{\shortstack{1.58b 90\%}}} & {\rotatebox{90}{\shortstack{2 bits}}} & {\rotatebox{90}{\shortstack{3 bits}}} & {\rotatebox{90}{\shortstack{4 bits}}} & {\rotatebox{90}{\shortstack{5 bits}}} & {\rotatebox{90}{\shortstack{original}}} \\
{dataset} & {} & {} & {} & {} & {} & {} & {} & {} & {} \\
\midrule
Caco2 Wang & 0.673 \tiny (89\%) & 0.713 \tiny (95\%) & 0.720 \tiny (95\%) & 0.679 \tiny (90\%) & 0.690 \tiny (92\%) & 0.733 \tiny (97\%) & 0.745 \tiny (99\%) & \textbf{0.748 \tiny (99\%)} & \textbf{\underline{0.750 \tiny (100\%)}} \\
HydrationFreeEnergy FreeSolv & 0.905 \tiny (98\%) & 0.907 \tiny (99\%) & 0.906 \tiny (98\%) & 0.893 \tiny (97\%) & 0.910 \tiny (99\%) & 0.913 \tiny (99\%) & \textbf{0.915 \tiny (99\%)} & \textbf{\underline{0.916 \tiny (99\%)}} & \textbf{\underline{0.916 \tiny (100\%)}} \\
LD50 Zhu & 0.572 \tiny (83\%) & 0.611 \tiny (89\%) & 0.618 \tiny (90\%) & 0.533 \tiny (77\%) & 0.596 \tiny (87\%) & 0.636 \tiny (92\%) & 0.669 \tiny (97\%) & \textbf{0.679 \tiny (99\%)} & \textbf{\underline{0.685 \tiny (100\%)}} \\
Lipophilicity (az) & 0.658 \tiny (87\%) & 0.700 \tiny (92\%) & 0.705 \tiny (93\%) & 0.637 \tiny (84\%) & 0.669 \tiny (88\%) & 0.713 \tiny (94\%) & 0.741 \tiny (98\%) & \textbf{0.749 \tiny (99\%)} & \textbf{\underline{0.754 \tiny (100\%)}} \\
PPBR AZ & 0.561 \tiny (98\%) & 0.563 \tiny (98\%) & 0.564 \tiny (99\%) & 0.555 \tiny (97\%) & 0.565 \tiny (99\%) & 0.568 \tiny (99\%) & \textbf{0.569 \tiny (99\%)} & \textbf{\underline{0.570 \tiny (99\%)}} & \textbf{\underline{0.570 \tiny (100\%)}} \\
Solubility AqSolDB & 0.838 \tiny (94\%) & 0.853 \tiny (96\%) & 0.852 \tiny (96\%) & 0.756 \tiny (85\%) & 0.842 \tiny (94\%) & 0.864 \tiny (97\%) & 0.879 \tiny (99\%) & \textbf{0.884 \tiny (99\%)} & \textbf{\underline{0.887 \tiny (100\%)}} \\
VDss Lombardo & 0.570 \tiny (99\%) & 0.572 \tiny (99\%) & 0.572 \tiny (99\%) & 0.560 \tiny (97\%) & 0.572 \tiny (99\%) & 0.573 \tiny (99\%) & \textbf{0.575 \tiny (99\%)} & \textbf{0.575 \tiny (99\%)} & \textbf{\underline{0.576 \tiny (100\%)}} \\
\end{tabular}